\documentclass[letterpaper,journal]{IEEEtran}

\usepackage{ieee_style}

\title{Learning to Probe LLM-Guided Controllers for Hallucinations via Conformally Regularized Prompt Sampling}

\author{
Sihang Wei, Melkior Ornik, and Hiroyasu Tsukamoto
\thanks{All authors are with the Department of Aerospace Engineering, The Grainger College of Engineering, University of Illinois Urbana-Champaign, Urbana, IL, 61801 USA. {\tt\small \{sihangw2, mornik, hiroyasu\}@illinois.edu}.
}%
}

\begin{document}
\maketitle
\begin{abstract}
Large language models (LLMs), when deployed in robot control, can produce syntactically valid yet task-objective-inconsistent closed-loop behavior. We study a performance predictor that sample-efficiently learns to detect such control-relevant hallucinations in LLM-guided model predictive control. Since closed-loop performance evaluations require costly simulation or hardware rollouts, we introduce conformal entropy-regularized excitation sampling (CERES), which selectively samples evaluations to enhance the predictor accuracy under a limited evaluation budget. CERES is designed to balance the potential information gained from each evaluation against prediction uncertainty that further learning is unlikely to reduce.
As new evaluations are observed, least-squares adaptation updates the predictor online, with dynamically tuned adaptive conformal inference calibrating such uncertainty to maintain the target coverage. We establish
(i) prediction regret bounds with an explicit condition for improvement over the existing worst-case bounds,
(ii)~local prediction coverage and entropy bounds for controller performance, and 
(iii)~a characterization of the tradeoff between feature exploration and uncertainty aversion under a finite evaluation budget. 
Experiments on nonlinear control tasks span four robotic platforms and include parallel evaluation of compatible robot-prompt pairs. In the multi-robot experiment, CERES achieves a 9\% reduction in mean squared prediction error and a 60\% relative increase in $R^2$ over the offline predictor in 9 and 8 batch rounds, respectively. The results are available at \myhref{https://ceres-robotics.netlify.app/}{\textbf{https://ceres-robotics.netlify.app/}}.
\end{abstract}


\section{Introduction}
\label{sec:introduction}
\IEEEPARstart{L}{arge} language models (LLMs) have shown growing capabilities to connect natural language task specifications with robotic decision-making across different levels of abstraction~\cite{firoozi2025foundation,yao2026bridging,bai2026embodied,catalano2026scene}.
In safety- and performance-critical robotic applications, however, controllers generated with LLMs cannot be reliably deployed without subsequent verification.
\begin{figure}
    \centering
    \includegraphics[width=0.4\textwidth]{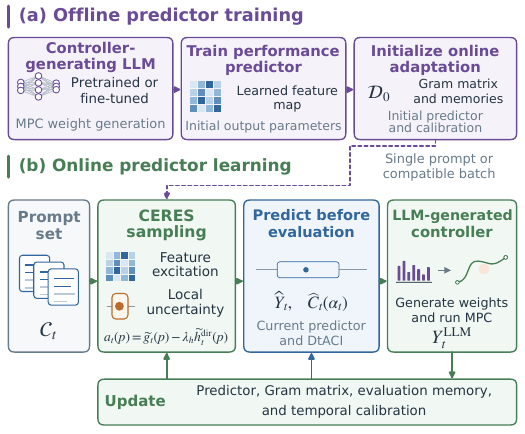}
    \caption{Graphical overview of CERES.}
    \label{fig:ceres_framework}
\end{figure}
\subsubsection*{\textbf{Background and Motivation}}
Establishing such reliability for general settings with general performance metrics is indeed a grand challenge, but even formal verification of LLM-generated parameters used to configure downstream controllers remains insufficiently understood. As a representative starting point, we consider model predictive control (MPC) as the downstream controller, where an LLM parametrically guides the MPC objective given a compact description of a robot control problem and user-specified performance priorities.

Prior studies have similarly explored how LLMs can translate language-driven task specifications into robot control, e.g., in MPC reward design~\cite{yu2023language}, reward refinement via reinforcement learning~\cite{ma2024eureka}, generation of task success criteria~\cite{turcato2025autonomous}, formalization of semantic constraints~\cite{guo2025castl}, and synthesis of verifiable task plans with dynamic affordance reasoning~\cite{zhou2026ambiguous}, just to name a few. Yet even syntactically and semantically valid LLM responses here may lead to closed-loop robot behavior inconsistent with overarching task objectives, resulting in \emph{control-relevant hallucinations}. In this work, therefore, we aim to detect such hallucinations by learning to predict a controller performance score defined over the full course of task execution.
If learned accurately, the score reflects how well observed robot behavior aligns with specified task priorities \emph{in closed loop} without executing each controller.
\subsubsection*{\textbf{Related Work}}
A central difficulty, particularly in robotics, is that sampling performance evaluation data for learning such a predictor can be prohibitively expensive. Each evaluation may involve repeated LLM inference, MPC problem construction and optimization, and closed-loop execution throughout a complete rollout in high-fidelity simulations and/or physical robot experiments. Exhaustively evaluating a large set of candidate controller configurations is therefore impractical. 

Such an evaluation bottleneck is not unique to LLM-guided controllers and has motivated sample-efficient approaches to control-related parameter tuning. For example, to improve control policies with fewer costly evaluations, prior studies have used Bayesian optimization to prioritize informative or safety-constrained evaluations for tuning controller parameters~\cite{marco2016automatic,berkenkamp2016safe}; combined simulated and physical robot evaluations to model policy objectives with fewer hardware experiments~\cite{he2025simulation}; and identified causally relevant configuration options through simulation to reduce the optimization search space~\cite{hossen2025cure}. Prior studies have developed learned predictors of language-model performance and robot task success. Schellaert et al.~\cite{schellaert2025analysing} train assessors to predict a language model's success on individual inputs. Learned value functions estimate the likelihood of successful robot skill execution from the current state~\cite{ichter2023saycan}. Latent world models have also been combined with vision-language evaluators
to predict the outcomes of proposed robot action
plans and assess whether those outcomes satisfy
task instructions~\cite{wu2025forewarn}. Our focus is to predict the closed-loop performance score of an LLM-generated controller from its task prompt before execution, and to improve this predictor using a limited number of selectively acquired controller evaluations.

Learning this predictor with limited evaluation budget requires both online adaptation and informative prompt selection. Online convex optimization provides regret bounds for cumulative prediction loss~\cite{hazan2016introduction}, while active learning,
core-set selection, and linear stochastic bandits
motivate selecting evaluations using prediction
uncertainty and feature information~\cite{cohn1996active,sener2018active,
abbasi2011improved}. As the evaluated tasks and predictor
evolve, prediction error statistics may change over
time, motivating temporal uncertainty calibration.
Conformal inference enables uncertainty calibration using nonconformity scores computed from finite samples without knowing their distribution~\cite{vovk2005algorithmic,angelopoulos2023conformal}. To accommodate changing error statistics, adaptive conformal inference (ACI) adjusts the calibration rate under continual distribution shifts~\cite{gibbs2021adaptive}, and dynamically tuned ACI (DtACI) further adapts the update gain~\cite{JMLR:v25:22-1218}. Control applications include constraint tightening with learned dynamics~\cite{chee2024uncertainty}, temporal logic control~\cite{yu2026signal}, probabilistic safety and stability certificates~\cite{hsu2025statistical}, and tracking error bounds via conformal contraction~\cite{wei2025conformal}. In robotics, conformal methods quantify uncertainty in perception~\cite{doula2025conformal} and teleoperation~\cite{zhao2024conformalized}, learn safety warnings
from human feedback~\cite{feldman2026learning}, and guide requests
for assistance~\cite{ren2023robots} and navigation among pedestrians~\cite{yao2025generalizable}.


While such uncertainty calibration techniques can account for changing prediction errors online, their uncertainty bound is shared across candidate task prompts and therefore does not capture their local differences. The feature-based exploration discussed earlier for online learning can favor poorly represented prompt features, but does not account for discrepancies between predicted and realized performance of LLM-guided controllers. These discrepancies may arise from either systematic prediction bias that further evaluations can reduce or performance variability that remains unpredictable from the learned features. Directly penalizing their magnitude can thus discourage evaluations that would help improve the predictor. This motivates combining feature excitation with local uncertainty calibrated after compensating for prediction bias, enabling more informative evaluations for learning to identify control-relevant hallucinations.
\subsubsection*{\textbf{Contributions}}
This paper introduces \emph{conformal entropy-regularized excitation sampling} (CERES), which selectively evaluates task prompts to sample-efficiently predict the closed-loop performance of LLM-guided controllers online. After each evaluation, the observed performance score updates the predictor's output parameters through online least squares, and DtACI adjusts temporal prediction intervals using the online-sampled sequence of prediction errors. To determine which task prompt to evaluate next, CERES rewards candidates that excite poorly represented directions in prompt-feature space while simultaneously penalizing uncertainty that remains after correcting systematic prediction bias.

Specifically, for each candidate prompt, an affine model fitted to its $K$~nearest previously evaluated prompts bounds the local bias between predicted and observed performance scores. The absolute errors after bias correction serve as local nonconformity scores, whose quantile determines the width of a candidate-specific prediction interval.
CERES uses the logarithm of this width to regularize feature excitation, prioritizing evaluations that improve the performance predictor without treating correctable prediction bias as persistent uncertainty. The contributions of this work are summarized as follows:
\begin{enumerate}
    \item We derive an expected static prediction regret bound for online adaptation of our predictor to identify an explicit condition under which the bound is strictly tighter than the existing worst-case bounds.
    \item We bound the probability that local conformal inference sets contain the true performance score despite imperfect bias correction, neighboring-prompt differences, and quantiles computed from limited data, relating their logarithmic size to conditional entropy to justify entropy regularization in CERES.
    \item We quantify the feature exploration--uncertainty aversion tradeoff in CERES, proving that the selected prompt retains a guaranteed fraction of maximum feature excitation with no wider local conformal inference set than excitation-only prompt selection.
    \item We extend CERES to parallel evaluations across heterogeneous robotic platforms, achieving a $9\%$ reduction in held-out mean squared prediction error and a $60\%$ relative increase in $R^2$ over the offline predictor in $9$ and $8$ batch rounds, respectively, requiring fewer rounds than any compared baseline.
\end{enumerate}
Note that \textbf{our problem formulation can accommodate broader control schemes beyond MPC}, such as reinforcement learning and classical feedback control, provided they admit tunable design parameters and performance evaluation.

The remainder of the paper introduces the technical problem formulation with preliminaries on online learning and uncertainty quantification (Section~\ref{sec:preliminaries}), develops CERES and its theoretical guarantees (Section~\ref{sec:guaranteed_sampling}), and then presents the experimental methodology (Section~\ref{sec:experimental_methodology}), results (Section~\ref{sec:result}), and discussion (Section~\ref{sec:conclusion}). Videos and experimental results are also available at
\myhref{https://ceres-robotics.netlify.app/}{\textbf{https://ceres-robotics.netlify.app/}}.
\section{Preliminaries and Problem Statement}
\label{sec:preliminaries}
We consider a family $\mathcal I$ of nonlinear robotic control tasks described in natural language. Each task $I\in\mathcal I$ specifies a discrete-time dynamics map $F_I$, an initial state $x_0$, admissible state and input sets $\mathcal X_I$ and $\mathcal U_I$, a target set $\mathcal X_{T,I}$, and a task horizon $N_I\in\mathbb N$. The system evolves according to
\begin{equation}
x_{k+1}=F_I(x_k,u_k),
~ k=0,\ldots,N_I-1,
\label{eq:task_dynamics}
\end{equation}
where $x_k$ and $u_k$ are the state and control input at step $k$. The target set is defined by $\mathcal X_{T,I}\coloneqq\{x\in\mathcal X_I:\phi_{T,I}(x)\leq0\}$ for a scalar function $\phi_{T,I}$. The task family accommodates systems with different state and input dimensions.  Executing a controller on task $I$ produces closed-loop state and input trajectories $X\coloneqq(x_0,\ldots,x_{N_I})$ and $U\coloneqq(u_0,\ldots,u_{N_I-1})$.

To evaluate the resulting behavior, let $j\in\mathcal J$ index a finite set of performance attributes, with normalized scores $S_j(X,U;I)\in[0,1]$ and preference profile $\rho\coloneqq(\beta_j)_{j\in\mathcal J}$ with $\beta_j\geq0$ and $\sum_{j\in\mathcal J}\beta_j>0$. The closed-loop performance score is defined as
\begin{equation}
Y(X,U;I,\rho)
\coloneqq
\tfrac{
\sum_{j\in\mathcal J}\beta_j S_j(X,U;I)
}{
\sum_{j\in\mathcal J}\beta_j
}
\in[0,1].
\label{eq:prelim-performance-score}
\end{equation}
Higher scores indicate better performance under the specified preference profile. Our experiments in Section~\ref{sec:experimental_methodology} consider objective completeness, arrival time, control effort, and control smoothness, with their scores defined in Appendix~\ref{app:score}.
\subsection{Closed-Loop Evaluation of LLM-Guided Controllers}
\label{sec:llm_controller_evaluator}
Given a task~$I$ and performance preference profile $\rho$, a structured prompt $p\coloneqq p(I,\rho)\in\mathcal S_p$ requests model predictive control~(MPC) cost weights from a large language model~(LLM).
For a prompt $p$, let $P_\theta(\cdot|p)$ denote the LLM response distribution with parameters $\theta$, and let $a\coloneqq M_\theta(p)$ denote its decoded response. The validity indicator $\nu^{\mathrm{LLM}}(a)\in\{0,1\}$ equals one if $a$ has the required format and yields MPC cost weights of the prescribed dimension with finite, nonnegative entries, and zero otherwise. For a valid response, the MPC parameterized by $w$ produces closed-loop trajectories $(X^p,U^p)$ on task $I$. Using the score in Eq.~\eqref{eq:prelim-performance-score}, we define the controller performance score as
\begin{equation}
    Y^{\mathrm{LLM}}(p)
    \coloneqq
    \begin{cases}
        Y(X^p,U^p;I,\rho), & \nu^{\mathrm{LLM}}(a)=1,\\
        0, & \nu^{\mathrm{LLM}}(a)=0.
    \end{cases}
    \label{eq:llm_controller_performance}
\end{equation}
We refer to invalid responses and valid responses yielding poor closed-loop performance as \emph{control-relevant hallucinations}.

Getting $Y^{\mathrm{LLM}}(p)$ requires LLM inference, MPC optimization, and a closed-loop rollout, making each evaluation costly. We consider adaptive prompt selection under a finite evaluation budget, together with online adaptation of a performance predictor with a fixed feature representation and an affine output layer. When available, previously collected prompt--performance pairs $(p,Y^{\mathrm{LLM}}(p))$ may provide an offline-trained warm start, although such training is not required.
\subsubsection{Online Performance Prediction}
\label{sec:online_prediction_preliminaries}
Let $\Phi:\mathcal S_p\rightarrow\mathbb R^{d_\phi+1}$ denote the feature map obtained by augmenting the $d_\phi$-dimensional representation with a constant~$1$ to account for the output-layer bias, and let $\omega\in\mathbb R^{d_\phi+1}$ contain the output weights and bias. The predicted performance is computed nominally by $\widehat Y(p;\omega)\coloneqq\Phi(p)^\top\omega$. During online adaptation, $\Phi$ remains fixed while only $\omega$ is updated.

At evaluation round $t$, the performance of the selected prompt $p_t$ is predicted as $\widehat Y_t\coloneqq\Phi_t^\top\widehat\omega_t$, where $\Phi_t\coloneqq\Phi(p_t)$ and $\widehat\omega_t$ is the current output-layer parameter estimate. After observing the actual performance $Y_t^{\mathrm{LLM}}\coloneqq Y^{\mathrm{LLM}}(p_t)$, the prediction residual is
\begin{equation}
    \xi_t
    \coloneqq Y_t^{\mathrm{LLM}}-\widehat Y_t
    =Y_t^{\mathrm{LLM}}-\Phi_t^\top\widehat\omega_t.
    \label{eq:online-prediction-residual}
\end{equation}
To update the output layer online, we use the squared prediction loss $\ell_t(\omega)\coloneqq(Y_t^{\mathrm{LLM}}-\Phi_t^\top\omega)^2$, for which $\ell_t(\widehat\omega_t)=\xi_t^2$ and $\nabla_\omega\ell_t(\widehat\omega_t)=-2\Phi_t\xi_t$. Let $\Omega\subset\mathbb R^{d_\phi+1}$ be a nonempty, compact, and convex parameter set, and let $\Pi_\Omega$ denote Euclidean projection onto $\Omega$. Starting from the offline-trained parameter $\widehat\omega_1=\omega^{\mathrm{off}}\in\Omega$, projected online gradient descent gives
\begin{equation}
    \widehat\omega_{t+1}
    =
    \Pi_\Omega\!\left(
        \widehat\omega_t
        -\eta_t\nabla_\omega\ell_t(\widehat\omega_t)
    \right)
    =
    \Pi_\Omega\!\left(
        \widehat\omega_t
        +2\eta_t\Phi_t\xi_t
    \right),
    \label{eq:online-output-layer-update-squared}
\end{equation}
where $\eta_t>0$ is the learning rate. To evaluate prediction performance, we compare the cumulative loss against the best fixed parameter in hindsight, $\omega^\star\in\arg\min_{\omega\in\Omega}\sum_{t=1}^{T}\ell_t(\omega)$. The resulting static prediction regret is
\begin{equation}
    \operatorname{Regret}_T
    \coloneqq
    \sum_{t=1}^{T}
    \left[
        \ell_t(\widehat\omega_t)
        -
        \ell_t(\omega^\star)
    \right].
    \label{eq:prediction_regret_background}
\end{equation}
This measures the cumulative prediction loss relative to the best fixed parameter on the same evaluation sequence.
\subsubsection{Online Performance Calibration}
\label{sec:adaptive_conformal_preliminaries}
In addition to updating the predictor online, the residuals in Eq.~\eqref{eq:online-prediction-residual} can be used to calibrate prediction intervals through adaptive conformal inference (ACI), with their magnitudes $s_t\coloneqq|\xi_t|$ serving as nonconformity scores.

Let $r\in\mathbb N$ be the calibration window length and $\mathcal W_1^r$ a nonempty initial sequence of at most $r$ nonconformity scores. Before round $t$, the window is $\mathcal W_t^r\coloneqq\operatorname{Last}_r(\mathcal W_1^r\mathbin{\Vert}(s_1,\ldots,s_{t-1}))$, where $\mathbin{\Vert}$ denotes concatenation and $\operatorname{Last}_r$ retains the most recent $r$ entries, or all entries if fewer are available. For a window $\mathcal W$ containing $n$ scores, let $Q_u(\mathcal W)$ denote the empirical $u$-quantile, computed by linear interpolation of the sorted scores at position $1+(n-1)u$. The prediction interval at calibration rate $a\in[0,1]$ is
\begin{equation}
    \widehat C_t(a)
    \coloneqq
    [
        \widehat Y_t-q_t(a),
        \widehat Y_t+q_t(a)
    ],
    \label{eq:global_conformal_interval}
\end{equation}
where $q_t(a)\coloneqq Q_{1-a}(\mathcal W_t^r)$. To target miscoverage rate $\alpha\in(0,1)$, ACI adjusts the calibration rate $\alpha_t\in[0,1]$ using observed coverage. After observing $Y_t^{\mathrm{LLM}}$, define
\begin{equation}
    \beta_t
    \coloneqq
    \sup
    \big\{
        a\in[0,1]:
        Y_t^{\mathrm{LLM}}\in\widehat C_t(a)
    \big\},
    \label{eq:dtaci-beta}
\end{equation}
with $\beta_t=0$ if the set is empty. The pinball loss $\ell_\alpha(a;\beta)\coloneqq\alpha(\beta-a)-\min\{0,\beta-a\}$ yields the projected update
\begin{equation}
\begin{aligned}
    \alpha_{t+1}
    &=
    \Pi_{[0,1]}
    [
        \alpha_t
        -
        \gamma
        \left.
        \partial_a\ell_\alpha(a;\beta_t)
        \right|_{a=\alpha_t}
    ] \\
    &=
    \Pi_{[0,1]}
    [
        \alpha_t
        +
        \gamma
        \left(
            \alpha-
            \mathbf 1\{\beta_t<\alpha_t\}
        \right)
    ],
\end{aligned}
    \label{eq:aci-alpha-update}
\end{equation}
where $\gamma>0$ is the adaptation gain, $\partial_a\ell_\alpha$ denotes a subgradient, and $\Pi_{[0,1]}$ denotes Euclidean projection onto $[0,1]$~\cite{gibbs2021adaptive}.

To reduce sensitivity to $\gamma$, dynamically tuned ACI (DtACI) maintains $N_{\mathrm e}$ experts with distinct gains $\{\gamma_j\}_{j=1}^{N_{\mathrm e}}$~\cite{JMLR:v25:22-1218}. Each follows Eq.~\eqref{eq:aci-alpha-update}, with selection weights updated exponentially based on pinball losses. At each round, DtACI samples an expert according to these weights and deploys its calibration rate as $\alpha_t$. Example settings are given in Section~\ref{sec:online_evaluation_metric}.
\subsection{CERES Problem Statement}
\label{sec:selective_evaluation_problem}
We formulate our problem of conformally regularized excitation sampling (CERES) as follows.
\begin{problem}
\label{prob:online_llm_performance_learning}
Consider a budget of $T$ controller evaluations, with a finite set of feasible candidate prompts $\mathcal C_t\subseteq\mathcal S_p$ at each round $t$. Given the initial predictor $(\Phi,\widehat\omega_1)$ and any previously collected prompt--performance pairs, each prompt $p_t\in\mathcal C_t$ must be selected using only the candidate sets available up to round $t$ and observations $(p_i,Y_i^{\mathrm{LLM}})$ from preceding rounds $i<t$, before observing $Y_t^{\mathrm{LLM}}$. Design online prediction, uncertainty calibration, and adaptive prompt sampling procedures to achieve the following technical objectives:
\begin{enumerate}
    \item \label{obj_1}
    \textbf{Online performance prediction.}
    Establish a finite-horizon bound on the expected static prediction regret in Eq.~\eqref{eq:prediction_regret_background} using calibrated residual quantiles and miscoverage, and identify a condition for strict improvement over the existing worst-case residual bound .

    \item \label{obj_2}
    \textbf{Local prediction uncertainty.}
    Estimate uncertainty for candidate prompts using neighboring evaluations in feature space. Derive local coverage bounds accounting for bias estimation, neighborhood approximation, and finite-sample quantile estimation errors, and bound the conditional differential entropy of the performance score in terms of the local prediction interval width.

    \item \label{obj_3}
    \textbf{Adaptive prompt sampling.}
    Select prompts using feature excitation and estimated local uncertainty. Quantify their tradeoff and the maximum excitation remaining after a finite number of evaluations, and assess prediction accuracy under the evaluation budget.
\end{enumerate}
\end{problem}
\section{Online Prediction Guarantee and Entropy-Regularized Sampling}
\label{sec:guaranteed_sampling}
This section presents the theoretical analysis of CERES. The theoretical results are organized as follows.
\begin{itemize}
\item\textit{\textcolor{uiucblue}{Results on  Objective~\ref{obj_1}}}:
    Section~\ref{sec:calibrated_prediction_regret} uses a DtACI coverage bound in Lemma~\ref{prop:dtaci-local-calibration} to derive Theorem~\ref{thm:online-ls-regret} on the expected static prediction regret bound.
\item\textit{\textcolor{uiucblue}{Results on Objective~\ref{obj_2}}}:
    Section~\ref{sec:local_conformal_analysis} estimates local prediction bias to construct uncertainty
    statistics from the compensated residuals, which yield the local coverage bound in Lemma~\ref{prop:local_knn_coverage} and the conditional differential entropy bound in Proposition~\ref{prop:local_conformal_entropy_bound}.
\item\textit{\textcolor{uiucblue}{Results on  Objective~\ref{obj_3}}}:
    Section~\ref{sec:feature_excitation} introduces the feature excitation criterion, which Section~\ref{sec:entropy_regularized_sampling} combines with local uncertainty in the CERES sampling rule of Algorithm~\ref{alg:online-sampling-adaptation}. Section~\ref{sec:ceres_sampling_guarantee} establishes the excitation--uncertainty tradeoff and finite-budget feature coverage bound in Theorem~\ref{thm:ceres_sampling_efficiency}.
\end{itemize}
Table~\ref{tab:analytical_roadmap} also summarizes the correspondence between these objectives and the supporting theoretical results.
\begin{table*}[htbp]
\centering
\caption{Correspondence between the problem objectives,
method components, and theoretical results.}
\label{tab:analytical_roadmap}
\scriptsize
\setlength{\tabcolsep}{5pt}
\renewcommand{\arraystretch}{1.18}
\renewcommand{\tabularxcolumn}[1]{m{#1}}

\begin{tabularx}{\textwidth}{
    @{}
    >{\raggedright\arraybackslash}m{0.21\textwidth}
    >{\raggedright\arraybackslash}m{0.21\textwidth}
    >{\raggedright\arraybackslash}X
    @{}
}
\toprule
\textbf{Problem objective}
&
\textbf{Method component}
&
\textbf{Supporting result and role}
\\
\midrule

Objective~\ref{obj_1}: Online performance prediction
&
Projected online least squares with DtACI residual calibration
&
Lemma~\ref{prop:dtaci-local-calibration} supplies the miscoverage bound used in Theorem~\ref{thm:online-ls-regret}. The theorem refines the expected static prediction regret bound and gives a condition for strict improvement over existing bounds.
\\
\addlinespace

Objective~\ref{obj_2}: Local prediction uncertainty
&
Local affine residual estimation and neighborhood quantiles
&
Lemmas~\ref{lem:directional_residual_separation} and~\ref{lem:local_quantile_error} bound residual separation
and quantile estimation errors. Lemma~\ref{prop:local_knn_coverage} and Proposition~\ref{prop:local_conformal_entropy_bound} use them to establish local coverage and a conditional differential entropy bound.
\\
\addlinespace

Objective~\ref{obj_3}: Adaptive prompt sampling
&
Feature excitation and local uncertainty in CERES
&
Theorem~\ref{thm:ceres_sampling_efficiency} establishes the excitation uncertainty
tradeoff at each round, and bounds the maximum excitation remaining after finitely many evaluations.
\\
\bottomrule
\end{tabularx}
\end{table*}
\subsection{Online Prediction Regret and Feature Excitation}
\label{sec:calibrated_prediction_regret}
Let $\mathcal H_{t-1}\coloneqq(\Phi,\widehat\omega_1,\mathcal D_0,(\mathcal C_i)_{i=1}^{t},(p_i,Y_i^{\mathrm{LLM}})_{i=1}^{t-1})$ denote the information available before selecting $p_t$. Here, $\Phi$ is the prompt feature map, $\widehat\omega_1$ denotes the initial predictor parameters, $\mathcal D_0$ is the initial dataset of offline-evaluated prompt--performance pairs, $\mathcal C_i$ is the candidate prompt set at round $i$, and $(p_i,Y_i^{\mathrm{LLM}})$ denotes the online-evaluated prompt and its observed performance score at round $i$. We define an $\mathcal H_{t-1}$-measurable oracle nominal miscoverage rate $\alpha_t^\star\in[0,1]$ for the conformal prediction interval, satisfying $\mathbb P(\beta_t<\alpha_t^\star\mid\mathcal H_{t-1})=\alpha$ for $\beta_t$ given in~\eqref{eq:dtaci-beta}, with $\alpha$ being the target miscoverage rate. The notation here largely follows that of Section~\ref{sec:llm_controller_evaluator}. Note that the oracle rate $\alpha_t^\star$ is the nominal miscoverage that would achieve the target $\alpha$ at round $t$. Since this rate may change as new observations become available, we measure its expected cumulative variation $V_{\mathcal T}^{\alpha}$ over $\mathcal T\coloneqq\{t_0,\ldots,t_1\}$ by $V_{\mathcal T}^{\alpha} \coloneqq \sum_{t=t_0+1}^{t_1} \mathbb E \left[|\alpha_t^\star-\alpha_{t-1}^\star|\right]$, where $t_0$ and $t_1$ denote the first and last round indices..

\begin{assumption}[Prediction and Temporal Calibration]
\label{ass:prediction_calibration}
Over rounds $t=1,\ldots,T$, the following conditions hold.
\begin{enumerate}
    \item The parameter set
    $\Omega\subset\mathbb R^{d_\phi+1}$ is nonempty,
    compact, and convex, with
    $\widehat\omega_1\in\Omega$.
    The feature vectors satisfy
    $\|\Phi_t\|_2\leq L_\Phi$ for some $L_\Phi>0$.

    \item The conditional density of $\beta_t$ given
    $\mathcal H_{t-1}$ is bounded below by $p_\beta>0$
    on $[0,1]$. The DtACI parameters satisfy the
    conditions in
    \cite[Sections~3.1--3.2]{JMLR:v25:22-1218}.

    \item For the target miscoverage rate
    $\alpha\in(0,1)$, the conditional coverage map
    \begin{equation}
        c_t^{\mathrm{cov}}(a;\mathcal H_{t-1})
        =
        \mathbb P\!\left(
            Y_t^{\mathrm{LLM}}\in\widehat C_t(a)
            \,\middle|\,\mathcal H_{t-1}
        \right)
        \label{eq:conditional_coverage_map}
    \end{equation}
    is $L_{\mathrm{cov}}$-Lipschitz in $a\in[0,1]$.
    The oracle rate satisfies
    $c_t^{\mathrm{cov}}(\alpha_t^\star;
    \mathcal H_{t-1})=1-\alpha$.
    The DtACI expert selection is conditionally
    independent of the current controller evaluation
    given $\mathcal H_{t-1}$.
\end{enumerate}
\end{assumption}
\subsubsection{DtACI Coverage}
Under Assumption~\ref{ass:prediction_calibration},
the DtACI tracking bound over consecutive evaluation rounds $\mathcal T=\{t_0,\ldots,t_1\}$ is
\begin{equation}
    \tfrac{1}{|\mathcal T|}
    \sum_{t\in\mathcal T}
    \tfrac{p_\beta}{2}
    \mathbb E
    \big[
        \left(\alpha_t-\alpha_t^\star\right)^2
    \big]
    \leq
    B_{\mathcal T}.
    \label{eq:dtaci-local-alpha-bound}
\end{equation}
Here, $B_{\mathcal T}$ denotes the finite-grid bound obtained by combining the DtACI aggregation inequality in~\cite[Lemma 2]{JMLR:v25:22-1218} with the projected-expert regret bound in Appendix~\ref{app:proof_dtaci_local_calibration}. The bound depends on the interval length and the variation of the oracle rate. The following lemma uses this tracking bound to quantify deviation from the target coverage probability.
\begin{lemma}[DtACI Coverage Bound]
\label{prop:dtaci-local-calibration}
Suppose that Assumption~\ref{ass:prediction_calibration}
holds and Eq.~\eqref{eq:dtaci-local-alpha-bound} applies over $\mathcal T=\{t_0,\ldots,t_1\}$. With
$\varepsilon_{\mathcal T}=L_{\mathrm{cov}}\sqrt{2B_{\mathcal T}/p_\beta}$, the cumulative miscoverage probability satisfies
\begin{equation}
    \sum_{t\in\mathcal T}
    \mathbb P\!\left(
        Y_t^{\mathrm{LLM}}
        \notin\widehat C_t(\alpha_t)
    \right)
    \leq
    |\mathcal T|(\alpha+\varepsilon_{\mathcal T}).
    \label{eq:dtaci-summed-miscoverage}
\end{equation}
\end{lemma}
The proof is given in Appendix~\ref{app:proof_dtaci_local_calibration} and follows from the DtACI tracking inequality and the Lipschitz property of the conditional coverage map. The cumulative miscoverage bound in Eq.~\eqref{eq:dtaci-summed-miscoverage} is used below to bound the contribution of prediction errors outside the calibrated intervals.
\subsubsection{Prediction Regret}
We next derive a calibration-dependent prediction regret bound for Objective~\ref{obj_1} of the problem in Section~\ref{prob:online_llm_performance_learning}. The regret in Eq.~\eqref{eq:prediction_regret_background} compares the online predictor with the best fixed output-layer parameter $\omega^\star\in\Omega$ over the evaluated prompts. Here, $\Omega$ is the convex set used in the projected update in Eq.~\eqref{eq:online-output-layer-update-squared}. For notational simplicity in the following, we define $R_\Omega \coloneqq \sup_{\omega,\omega'\in\Omega} \|\omega-\omega'\|_2$ and $R_\omega \coloneqq \sup_{\omega\in\Omega}\|\omega\|_2$. Under Assumption~\ref{ass:prediction_calibration},
the projected update ensures
$\widehat\omega_t\in\Omega$. Since $Y_t^{\mathrm{LLM}}\in[0,1]$, the prediction
residual satisfies
\begin{equation}
    |\xi_t|
    =
    \left|
        Y_t^{\mathrm{LLM}}
        -\Phi_t^\top\widehat\omega_t
    \right|
    \leq M_\xi \coloneqq
    1+L_\Phi R_\omega.
    \label{eq:uniform_prediction_residual_bound}
\end{equation}
Let $\mathcal E_t\coloneqq\{Y_t^{\mathrm{LLM}}\in\widehat C_t(\alpha_t)\}=\{|\xi_t|\leq q_t(\alpha_t)\}$ denote the coverage event for the interval in Eq.~\eqref{eq:global_conformal_interval}, and let $\mathcal E_t^c$ denote its complement.
When the interval contains the observed score, the residual is bounded by the calibrated quantile $q_t(\alpha_t)$; otherwise, the worst-case bound $M_\xi$ applies. Thus, we get
\begin{equation}
    \xi_t^2
    \leq
    q_t^2(\alpha_t)\mathbf 1_{\mathcal E_t}
    +
    M_\xi^2\mathbf 1_{\mathcal E_t^c}.
    \label{eq:calibrated_residual_decomposition}
\end{equation}
Define the average squared residual quantile as
\begin{equation}
    \bar q_T^2
    \coloneqq
    \tfrac{1}{T}
    \sum_{t=1}^{T}
    \mathbb E \left[
        q_t^2(\alpha_t)
    \right].
    \label{eq:maximum_rms_quantile}
\end{equation}
Applying Lemma~\ref{prop:dtaci-local-calibration} with
$\mathcal T=\{1,\ldots,T\}$ and
$\varepsilon_T=\varepsilon_{\mathcal T}$ gives
$\sum_{t=1}^{T}\mathbb P(\mathcal E_t^c)
\leq T(\alpha+\varepsilon_T)$. Taking expectations in Eq.~\eqref{eq:calibrated_residual_decomposition} and summing over the horizon yields
\begin{equation}
    \sum_{t=1}^{T}\mathbb E[\xi_t^2]
    \leq
    T\left[
        \bar q_T^2
        +(\alpha+\varepsilon_T)M_\xi^2
    \right].
    \label{eq:calibrated_residual_energy_bound}
\end{equation}
The uniform residual bound in Eq.~\eqref{eq:uniform_prediction_residual_bound} gives $\sum_{t=1}^{T}\mathbb E[\xi_t^2]\leq TM_\xi^2$. The calibrated estimate in Eq.~\eqref{eq:calibrated_residual_energy_bound} is strictly smaller whenever $\bar q_T^2+(\alpha+\varepsilon_T)M_\xi^2<M_\xi^2$. Under this condition, substitution into the projected gradient regret inequality yields a strictly tighter regret bound for the same learning rate. The following theorem incorporates both residual bounds.
\begin{theorem}[Prediction Regret with Temporal Calibration]
\label{thm:online-ls-regret}
Suppose that
Assumption~\ref{ass:prediction_calibration}
holds and Lemma~\ref{prop:dtaci-local-calibration}
applies over rounds $1,\ldots,T$. Define the normalized residual bound
\begin{equation}
    \vartheta_T
    \coloneqq
    \min\left\{
        1,
        \tfrac{\bar q_T^2}{M_\xi^2}
        +\alpha+\varepsilon_T
    \right\}.
    \label{eq:calibrated_residual_factor}
\end{equation}
The projected update in
Eq.~\eqref{eq:online-output-layer-update-squared}
with a constant learning rate $\eta_t=\eta>0$ satisfies
\begin{equation}
    \mathbb E[\operatorname{Regret}_T]
    \leq
    \tfrac{R_\Omega^2}{2\eta}
    +2\eta L_\Phi^2 M_\xi^2 T\vartheta_T.
    \label{eq:expected-regret-final-theorem}
\end{equation}
If $R_\Omega>0$, choosing
$\eta=R_\Omega/(2L_\Phi M_\xi\sqrt T)$ yields
\begin{equation}
    \mathbb E[\operatorname{Regret}_T]
    \leq
    R_\Omega L_\Phi M_\xi\sqrt T\,
    (1+\vartheta_T).
    \label{eq:expected-regret-refined-rate-theorem}
\end{equation}
\end{theorem}
Theorem~\ref{thm:online-ls-regret} provides a calibration dependent refinement of the prediction
regret bound by incorporating residual quantiles and the cumulative miscoverage bound.
For the learning rate choice specified in the theorem, the upper bound in Eq.~\eqref{eq:expected-regret-refined-rate-theorem} is $(1+\vartheta_T)/2$ times the uniform residual bound $2R_\Omega L_\Phi M_\xi\sqrt T$. The improvement is strict whenever $\vartheta_T<1$. The proof appears in
Appendix~\ref{app:proof-online-ls-regret}.
\subsubsection{Feature Space Excitation}
\label{sec:feature_excitation}
Objective~\ref{obj_3} of the problem in Section~\ref{prob:online_llm_performance_learning} motivates sampling prompts that improve feature coverage for online output parameter estimation. We adopt a standard excitation criterion based on the regularized feature Gram matrix~\cite{abbasi2011improved}, which favors feature
directions weakly represented by previous evaluations.

At round $t$, we evaluate the excitation score over the candidate prompt set $\mathcal C_t$. Let $\{p_i^0\}_{i=1}^{N_0}$ denote the prompts in the initial dataset $\mathcal D_0$, with $N_0\coloneqq|\mathcal D_0|$. Using the frozen feature map $\Phi:\mathcal S_p\rightarrow\mathbb R^{d_\phi+1}$ introduced in Section~\ref{sec:online_prediction_preliminaries}, we define the regularized Gram matrix as
\begin{equation} 
    \Sigma_t \coloneqq \lambda_V I_{d_\phi+1} + \sum_{i=1}^{N_0} \Phi(p_i^0)\Phi(p_i^0)^\top + \sum_{s=1}^{t-1} \Phi(p_s)\Phi(p_s)^\top, 
\label{eq:gram_matrix_sampling} 
\end{equation} 
where $I_{d_\phi+1}\in \mathbb R^{(d_\phi+1)\times(d_\phi+1)}$ is the identity matrix and $\lambda_V>0$ ensures that $\Sigma_t$ is positive definite. The two sums contain the contributions from the initial dataset and the preceding online evaluations, respectively. Each new evaluation adds $\Phi(p_t)\Phi(p_t)^\top$ to form $\Sigma_{t+1}$. The feature excitation score for a candidate $p\in\mathcal C_t$ is
\begin{equation}
    g_t(p)\coloneqq
    \Phi(p)^\top\Sigma_t^{-1}\Phi(p)
    =
    \|\Phi(p)\|_{\Sigma_t^{-1}}^2.
    \label{eq:gram_excitation_sampling}
\end{equation}
Let $u_{t,\ell}$ and $\lambda_{t,\ell}>0$ denote the
orthonormal eigenvectors and corresponding eigenvalues
of $\Sigma_t$. Then $ g_t(p)= \sum_{\ell=1}^{d_\phi+1} \frac{(u_{t,\ell}^\top\Phi(p))^2}{\lambda_{t,\ell}}$
Feature components along directions with smaller eigenvalues receive larger weights, favoring directions less represented by previous evaluations. The criterion also admits a D-optimal experimental design interpretation. The matrix determinant lemma gives $\log\det ( \Sigma_t+\Phi(p)\Phi(p)^\top ) -\log\det(\Sigma_t) = \log \left(1+g_t(p)\right)$. Since $\log(1+z)$ is strictly increasing for $z\geq0$, maximizing $g_t$ maximizes the increase in
$\log\det\Sigma_t$ contributed by a single evaluation. Thus, the excitation score implements a sequential D-optimal design criterion in the learned feature space.

The issue here is that the excitation score depends only on feature vectors and does not incorporate prediction residuals. A large residual can arise from systematic prediction bias or from variation unexplained by the frozen feature representation. Additional evaluations may reduce systematic bias, so directly penalizing residual magnitude can discourage informative queries. The next subsection estimates local prediction bias from neighboring evaluations and constructs an uncertainty statistic from the bias-compensated residuals.
\subsection{Local Conformal Coverage and Entropy Bound}
\label{sec:local_conformal_analysis}

The temporal prediction interval in Eq.~\eqref{eq:global_conformal_interval}
uses the same residual quantile $q_t(\alpha_t)$ for every candidate prompt. Centering it at $\Phi(p)^\top\widehat\omega_t$ therefore gives prediction width $2q_t(\alpha_t)$ for every $p\in\mathcal C_t$. To address objective~\ref{obj_2}, we estimate local prediction uncertainty for each candidate prompt using previously evaluated prompts with similar feature representations.
Let $Y_t^{\mathrm{LLM}}(p)$ denote the performance score associated with evaluating prompt $p$ at round $t$.

\begin{assumption}[Sequential Sampling and Conditional Stationarity]
\label{ass:local_sampling_model}
The candidate prompt sets satisfy
$\mathcal C_t\subseteq\mathcal P$, where
$\mathcal P\subseteq\mathcal S_p$ is a finite prompt pool with $N_{\mathrm{pool}}=|\mathcal P|$. The pool also contains all prompts in the initialization dataset $\mathcal D_0$.
The pool $\mathcal P$ and feature map $\Phi$ are fixed before the evaluation sequence used in the local analysis. Each prompt is evaluated at most once, and all initialization
and preceding online observations are retained.

The index $i$ runs over the initialization evaluations followed by the online evaluations.
Let $\mathcal F_{i-1}$ denote the information available before evaluation $i$; prompt selection is causal. At online round $t$, $i=N_0+t$ and $\mathcal F_{i-1}=\mathcal H_{t-1}$. For each $p\in\mathcal P$,
$Y_i^{\mathrm{LLM}}(p)$ denotes its performance score if evaluated at index $i$.
There exists a fixed function $m_\Phi:\mathcal P\to[0,1]$ such that
\begin{equation}
    \mathbb E\!\left[
        Y_i^{\mathrm{LLM}}(p)
        \,\middle|\,\mathcal F_{i-1}
    \right]
    =
    m_\Phi(p),
    \qquad p\in\mathcal P,
    \label{eq:local_conditional_mean_assumption}
\end{equation}
throughout this evaluation sequence.
Moreover, $m_\Phi(p)=m_\Phi(p')$ whenever
$\Phi(p)=\Phi(p')$.

For each prompt $p$, its conditional noise-magnitude distribution $F_p^\varepsilon$ is independent of the evaluation index and preceding history. Along the selected sequence, with $\varepsilon_i=Y_i^{\mathrm{LLM}}-m_\Phi(p_i)$,
this gives
\begin{equation}
    \mathbb P\!\left(
        |\varepsilon_i|\leq u
        \,\middle|\,\mathcal F_{i-1},p_i
    \right)
    =
    F_{p_i}^{\varepsilon}(u).
    \label{eq:local_noise_stability}
\end{equation}
\end{assumption}


Under Assumption~\ref{ass:local_sampling_model},
$m_\Phi(p)$ is the conditional mean at online round $t$. The residual of the current predictor $\widehat Y_t(p)=\Phi(p)^\top\widehat\omega_t$
can therefore be decomposed as
\begin{equation}
    Y_t^{\mathrm{LLM}}(p)-\widehat Y_t(p)
    =
    \underbrace{
        m_\Phi(p)-\widehat Y_t(p)
    }_{b_t(p)}
    +
    \underbrace{
        Y_t^{\mathrm{LLM}}(p)-m_\Phi(p)
    }_{\varepsilon_t(p)},
\label{eq:local_residual_decomposition}
\end{equation}
where $b_t(p)$ is the conditional prediction bias and $\varepsilon_t(p)$ represents variation unexplained by the frozen feature representation, satisfying
$\mathbb E[\varepsilon_t(p)\mid\mathcal H_{t-1}]=0$. Since further evaluations may reduce the systematic bias (i.e., the first term), we estimate and compensate for it before measuring the remaining uncertainty.
\subsubsection{Local Prediction Bias Estimation}
To estimate the bias, we use previously evaluated prompts with similar features. Let $\mathcal I_t^{\mathrm{eval}}$ index the evaluations available before round $t$, including the initialization data. The local evaluation memory stores their feature vectors and performance scores as $\mathcal R_t \coloneqq \left\{ \bigl(\Phi(p_i),Y_i^{\mathrm{LLM}}\bigr) \right\}_{i\in\mathcal I_t^{\mathrm{eval}}}$.
For each stored evaluation, we compute the signed residual using the current predictor,
$\zeta_{i,t}\coloneqq
Y_i^{\mathrm{LLM}}-\Phi(p_i)^\top\widehat\omega_t$.
This allows the local bias estimate to reflect the current predictor rather than the predictor at the time of evaluation.

To define similarity, we perform neighbor selection based on the normalized feature representation $\bar\Phi(p)\coloneqq\Phi(p)/\|\Phi(p)\|_2$. Let $K$ denote the neighborhood size and set $K_t\coloneqq\min\{K,|\mathcal I_t^{\mathrm{eval}}|\}$. We select the $K_t$ previously evaluated prompts with the largest cosine similarities to candidate $p$. This $K$-nearest-neighbor (KNN) construction defines the index set
\begin{equation}
    \mathcal N_t^{K_t}(p)
    \coloneqq
    \operatorname{TopK}_{t,\,i\in\mathcal I_t^{\mathrm{eval}}}
    \left\{
        \bar\Phi(p)^\top\bar\Phi(p_i)
    \right\},
\label{eq:knn_neighbor_set}
\end{equation}
where $\operatorname{TopK}_t$ returns the corresponding indices. Distance ties are resolved by a prompt ordering fixed before the evaluations begin. 
The feature offset from candidate $p$ to a neighboring prompt $p_i$, and the corresponding neighborhood radius are
\begin{equation}
\begin{aligned}
    d_i(p)
    &\coloneqq
    \bar\Phi(p_i)-\bar\Phi(p),
    ~ i\in\mathcal N_t(p),\\
    \rho_t(p)
    &\coloneqq
    \max_{i\in\mathcal N_t(p)}
    \|d_i(p)\|_2.
\end{aligned}
\label{eq:local_directional_offsets}
\end{equation}

The prediction bias may vary even among neighboring prompts. Rather than treating it as constant, we approximate its local variation by an affine function of the feature offset $d_i(p)$. Its coefficients are estimated by regularized least squares:
\begin{equation}
\begin{aligned}
    (\widehat b_t(p),
    \widehat{\boldsymbol\beta}_t(p))
    \in
    \arg\min_{b,\boldsymbol\beta}\;
    &\sum_{i\in\mathcal N_t(p)}
    \left[
        \zeta_{i,t}
        -b
        -\boldsymbol\beta^\top d_i(p)
    \right]^2\\
    &+
    \lambda_{\mathrm{loc}}
    \|\boldsymbol\beta\|_2^2.
\end{aligned}
\label{eq:local_directional_fit}
\end{equation}
Here, $b\in\mathbb R$ and
$\boldsymbol\beta\in\mathbb R^{d_\phi+1}$ are optimization variables, and $\lambda_{\mathrm{loc}}>0$ is the regularization parameter. The intercept $\widehat b_t(p)$ estimates the prediction bias at candidate $p$, while
$\widehat{\boldsymbol\beta}_t(p)$ estimates its local gradient with respect to the normalized feature representation.
\subsubsection{Bias-Compensated Prediction Intervals}
We now use the fitted affine model to remove estimated systematic bias from each neighboring residual. The remaining absolute error defines the directional residual score:
\begin{equation}
    u_{i,t}^{\mathrm{dir}}(p)
    \coloneqq
    |
        \zeta_{i,t}
        -
        \widehat b_t(p)
        -
        \widehat{\boldsymbol\beta}_t(p)^\top d_i(p)
    |.
    \label{eq:directional_nonconformity}
\end{equation}
Subtracting the intercept and directional term compensates for estimated bias variation across the neighborhood. Let $\tau\in(0,1)$ denote the nominal local miscoverage rate. The empirical $(1-\tau)$-quantile of the directional scores is
\begin{equation}
    \widehat q_t^{\mathrm{dir}}(p;\tau)
    \coloneqq
    Q_{1-\tau}\!\left(
        \{u_{i,t}^{\mathrm{dir}}(p)\}_{i\in\mathcal N_t(p)}
    \right).
    \label{eq:knn_local_nonconformity_quantile}
\end{equation}

To construct the local interval, we first correct the current prediction using the estimated bias:
\begin{equation}
    \widehat m_t^{\mathrm{dir}}(p)
    \coloneqq
    \Pi_{[0,1]}
    [\widehat Y_t(p)+\widehat b_t(p)]
    ,
    \label{eq:directional_local_center}
\end{equation}
where $\Pi_{[0,1]}$ denotes projection onto the score range $[0,1]$. We then use the directional residual quantile to form
\begin{equation}
    \widehat C_t^{\mathrm{dir}}(p;\tau)
    \coloneqq
    [
        \widehat m_t^{\mathrm{dir}}(p)
        - \widehat q_t^{\mathrm{dir}}(p;\tau),
        \widehat m_t^{\mathrm{dir}}(p)
        +\widehat q_t^{\mathrm{dir}}(p;\tau)
    ]\cap[0,1].
    \label{eq:local_knn_sampling_interval}
\end{equation}
Thus, the interval center is corrected for estimated bias, while its width is determined by the remaining residual variation. Let
$V_t^{\mathrm{dir}}(p)\coloneqq
\mu(\widehat C_t^{\mathrm{dir}}(p;\tau))$
denote this width, where $\mu(\cdot)$ is Lebesgue measure. We define the local uncertainty penalty for prompt selection as
\begin{equation}
    h_t^{\mathrm{dir}}(p)
    \coloneqq
    \log V_t^{\mathrm{dir}}(p).
    \label{eq:local_conformal_sampling_term}
\end{equation}

\begin{figure}[t]
    \centering
    \includegraphics[width=0.95\columnwidth]
    {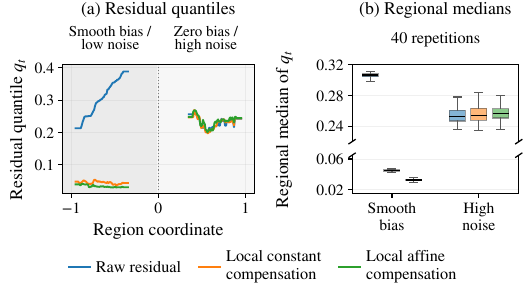}
    \caption{Local bias compensation in a synthetic example:
    residual quantiles (left) and regional medians over
    $40$ repetitions (right).}
    \label{fig:bias_noise_diagnostic}
\end{figure}

Figure~\ref{fig:bias_noise_diagnostic} illustrates the effect of bias compensation in a synthetic example.
One region has a smoothly varying prediction bias and small random variation, while the other has zero bias and larger random variation. In the first region, raw residual quantiles are dominated by systematic bias; local affine compensation yields a substantially smaller quantile than either raw residuals or constant compensation. All three estimates remain large in the high variation regions. The regional medians over $40$ repetitions show the same pattern.
\subsubsection{Local Coverage Guarantees}
The interval in Eq.~\eqref{eq:local_knn_sampling_interval} relies on estimates of both the conditional mean and the residual quantile. Its coverage can therefore be affected by three sources of error: imperfect local bias estimation, differences between neighboring noise distributions, and quantile estimation from finitely many evaluations. We next bound these errors and their effect on coverage. First, Lemma~\ref{lem:directional_residual_separation} relates the bias-compensated directional scores to the residual variation that remains under exact bias compensation.
\begin{lemma}[Directional Residual Separation]
\label{lem:directional_residual_separation}
Suppose that~$b_t$, expressed in normalized feature coordinates, has an $L_b$-Lipschitz gradient on a convex neighborhood containing $\bar\Phi(p)$ and $\{\bar\Phi(p_i):i\in\mathcal N_t(p)\}$. Suppose that the local regression estimates satisfy
\begin{equation}
    |\widehat b_t(p)-b_t(p)|
    +
    \rho_t(p)
    \|
        \widehat{\boldsymbol\beta}_t(p)
        -
        \nabla_{\bar\Phi}b_t(p)
    \|_2
    \leq
    \varepsilon_{\mathrm{reg},t}(p),
\label{eq:local_affine_fit_error}
\end{equation}
where $\nabla_{\bar\Phi}b_t(p)$ denotes the gradient with respect to the normalized feature coordinates. With $\varepsilon_i$ defined in Assumption~\ref{ass:local_sampling_model}, every $i\in\mathcal N_t(p)$ satisfies $| u_{i,t}^{\mathrm{dir}}(p)-|\varepsilon_i|| \leq \delta_t^{\mathrm{dir}}(p)$,
where $\delta_t^{\mathrm{dir}}(p) \coloneqq \varepsilon_{\mathrm{reg},t}(p) + \tfrac{L_b}{2}\rho_t(p)^2$.
\end{lemma}
The proof is given in Appendix~\ref{app:proof_directional_residual_separation}. The bound separates the local regression error $\varepsilon_{\mathrm{reg},t}(p)$ from the second-order error caused by variation in the bias across the neighborhood.

We next bound the error in estimating the candidate's noise quantile from neighboring evaluations. Let $F_{t,p}^{\varepsilon}(u) \coloneqq \mathbb P ( |\varepsilon_t(p)|\leq u \mid\mathcal H_{t-1})$ denote the conditional distribution function of the unexplained residual magnitude. Assumption~\ref{ass:local_sampling_model} gives
$F_{t,p}^{\varepsilon}=F_p^{\varepsilon}$. Its $(1-\tau)$-quantile is $q_{\varepsilon,t}^{\star}(p;\tau) \coloneqq \inf\{u:F_{t,p}^{\varepsilon}(u)\geq1-\tau\}$. 
The empirical quantile of the unobserved noise magnitudes is $ q_{\varepsilon,t}^{\mathrm{emp}}(p;\tau) \coloneqq Q_{1-\tau} \left( \{|\varepsilon_i|:i\in\mathcal N_t(p)\} \right)$.
These oracle quantities are used only in the analysis. To control the quantile estimation error under Assumption~\ref{ass:local_sampling_model},
we impose the following local regularity condition.
\begin{assumption}[Local Noise Quantile Regularity]
\label{ass:local_quantile_smoothness}
There exists $L_q>0$ such that $\left|q_{\varepsilon,t}^{\star}(p;\tau)- q_{\varepsilon,t}^{\star}(p';\tau) \right| \leq L_q\|\bar\Phi(p)-\bar\Phi(p')\|_2$
for all $p,p'\in\mathcal P$. There also exist constants $\underline f>0$ and $r_q>0$ such that each $F_p^{\varepsilon}$ admits a density $f_p^{\varepsilon}$ in a neighborhood of its oracle quantile, with
\begin{equation}
    f_p^{\varepsilon}(u)\geq\underline f
    ~\text{whenever}~
    \bigl|
        u-q_{\varepsilon,t}^{\star}(p;\tau)
    \bigr|\leq r_q.
    \label{eq:local_noise_density_lower_bound}
\end{equation}
\end{assumption}
Assumption~\ref{ass:local_sampling_model} allows adaptive prompt selection while keeping each prompt's noise distribution fixed over the local analysis. Assumption~\ref{ass:local_quantile_smoothness} limits differences in the noise quantiles of neighboring prompts and ensures that errors in their distribution estimates can be translated into quantile errors. These lead to the following lemma.
\begin{lemma}[Finite Sample Neighborhood Quantile Error]
\label{lem:local_quantile_error}
Suppose that Assumptions~\ref{ass:local_sampling_model} and~\ref{ass:local_quantile_smoothness} hold. Let $\delta_{\mathrm{conf}}\in(0,1)$ be a given failure probability. Define the statistical quantile error bound for $n\geq1$ observations by
\begin{equation}
    \varepsilon_{\mathrm{stat}}(n)
    \coloneqq
    \tfrac{1}{\underline f}
    \left[
        \sqrt{
            \tfrac{
                \log(2N_{\mathrm{pool}}^3/
                \delta_{\mathrm{conf}})
            }{2n}
        }
        +\tfrac{1}{n}
    \right].
    \label{eq:local_quantile_statistical_error}
\end{equation}
There exists an event $\mathcal E_{\mathrm{conf}}$ with $\mathbb P(\mathcal E_{\mathrm{conf}}) \geq1-\delta_{\mathrm{conf}}$ on which
\begin{equation}
\begin{aligned}
    \bigl|
        q_{\varepsilon,t}^{\mathrm{emp}}(p;\tau)
        -
        q_{\varepsilon,t}^{\star}(p;\tau)
    \bigr|
    \leq
    L_q\rho_t(p)
    +\varepsilon_{\mathrm{stat}}(K_t)
\end{aligned}
\label{eq:local_noise_quantile_concentration}
\end{equation}
holds simultaneously for every candidate $p\in\mathcal C_t$ at every round $t$ satisfying $K_t\geq1$ and $\varepsilon_{\mathrm{stat}}(K_t)\leq r_q$.
\end{lemma}
Appendix~\ref{app:proof_local_quantile_error} establishes this bound for adaptively selected evaluations. The first term, $L_q\rho_t(p)$, captures differences between the candidate's noise quantile and those of its neighbors. The second term, $\varepsilon_{\mathrm{stat}}(K_t)$, captures finite-sample estimation error. The condition $\varepsilon_{\mathrm{stat}}(K_t)\leq r_q$ keeps the statistical perturbation within the region where the density lower bound in Eq.~\eqref{eq:local_noise_density_lower_bound} applies.

We now combine these errors to bound coverage. To do so, let us additionally assume that the conditional noise cumulative distribution function (CDF) does not change too rapidly near its oracle quantile, limiting the loss of coverage caused by estimation errors.
\begin{assumption}[Local Conditional Distribution Regularity]
\label{ass:local_cdf_regularity}
Let $\Delta_t^{\mathrm{dir}}(p)$ denote the combined
error bound for the estimated conditional mean
and residual quantile:
\begin{equation}
    \Delta_t^{\mathrm{dir}}(p)
    \coloneqq
    \varepsilon_{\mathrm{reg},t}(p)
    +\delta_t^{\mathrm{dir}}(p)
    +
    L_q\rho_t(p)
    +\varepsilon_{\mathrm{stat}}(K_t).
\label{eq:directional_coverage_perturbation}
\end{equation}
There exists a constant $c>0$ such that, for every
candidate $p\in\mathcal C_t$ and every
$v\in[0,\Delta_t^{\mathrm{dir}}(p)]$, $F_{t,p}^{\varepsilon} \bigl(q_{\varepsilon,t}^{\star}(p;\tau)\bigr) - F_{t,p}^{\varepsilon} \bigl(q_{\varepsilon,t}^{\star}(p;\tau)-v\bigr) \leq cv$.
\end{assumption}
This assumption ensures that reducing the oracle noise threshold by $v$ decreases the corresponding coverage probability by at most $cv$.
\begin{lemma}[Local KNN Conformal Coverage]
\label{prop:local_knn_coverage}
Suppose that Assumptions~\ref{ass:local_sampling_model}, \ref{ass:local_quantile_smoothness}, and~\ref{ass:local_cdf_regularity} hold, together with the hypotheses of Lemma~\ref{lem:directional_residual_separation}. Define $\varepsilon_t^{\mathrm{dir}}(p) \coloneqq c\Delta_t^{\mathrm{dir}}(p)$. On the event $\mathcal E_{\mathrm{conf}}$ from
Lemma~\ref{lem:local_quantile_error}, every round satisfying $K_t\geq1$ and $\varepsilon_{\mathrm{stat}}(K_t)\leq r_q$ and every candidate $p\in\mathcal C_t$ satisfy
\begin{equation}
    \mathbb P (
        Y_t^{\mathrm{LLM}}(p)
        \in\widehat C_t^{\mathrm{dir}}(p;\tau)
        \mid\mathcal H_{t-1}
    )
    \geq
    1-\tau-\varepsilon_t^{\mathrm{dir}}(p).
    \label{eq:local_coverage_bound}
\end{equation}
\end{lemma}

Appendix~\ref{app:proof_local_knn_coverage} establishes this result by combining the directional residual error from Lemma~\ref{lem:directional_residual_separation} and the neighborhood quantile error from Lemma~\ref{lem:local_quantile_error}. The resulting coverage loss explicitly accounts for imperfect bias correction, differences between neighboring prompts, and finite-sample quantile estimation.
\subsubsection{Conditional Entropy Bound}
The coverage result above also provides an interpretation of the local interval width as an uncertainty measure. 
We finally use the coverage guarantee from Lemma~\ref{prop:local_knn_coverage} to relate the local interval width $V_t^{\mathrm{dir}}(p)$ to the conditional differential entropy of $Y_t^{\mathrm{LLM}}(p)$. This provides an information-theoretic interpretation of the interval width as a measure of uncertainty.

Since the performance score lies in the interval $[0,1]$, we follow the entropy decomposition underlying \cite[Proposition~3.2]{correia2024information}. Specifically, we partition the conditional distribution according to whether the performance score belongs to $\widehat C_t^{\mathrm{dir}}(p;\tau)$ or its complement. Combining this decomposition with the coverage guarantee yields the following entropy upper bound.
\begin{proposition}[Local KNN Conformal Entropy Bound]
\label{prop:local_conformal_entropy_bound}
Suppose that the conditions of Lemma~\ref{prop:local_knn_coverage} hold. Consider a round~$t$ satisfying $K_t\geq1$ and $\varepsilon_{\mathrm{stat}}(K_t)\leq r_q$.
Suppose that, for every $p\in\mathcal C_t$, the conditional distribution of $Y_t^{\mathrm{LLM}}(p)$ given $\mathcal H_{t-1}$ admits a density with respect
to Lebesgue measure on $[0,1]$, and that $V_t^{\mathrm{dir}}(p)>0$. On the event $\mathcal E_{\mathrm{conf}}$ defined in Lemma~\ref{lem:local_quantile_error}, let $\bar\eta_{\mathrm{dir}}\in(0,1/2)$ satisfy
\begin{equation}
    \bar\eta_{\mathrm{dir}}
    \geq
    \sup_{p\in\mathcal C_t}
    \left[
        \tau+\varepsilon_t^{\mathrm{dir}}(p)
    \right].
    \label{eq:uniform_local_eta_definition}
\end{equation}
Then, for every $p\in\mathcal C_t$,
\begin{equation}
    H\!\left(
        Y_t^{\mathrm{LLM}}(p)\mid\mathcal H_{t-1}
    \right)
    \leq
    h_b(\bar\eta_{\mathrm{dir}})
    +(1-\bar\eta_{\mathrm{dir}})
      h_t^{\mathrm{dir}}(p),
\label{eq:local_conformal_expected_volume_bound}
\end{equation}
where $H(\cdot|\mathcal H_{t-1})$ denotes the differential entropy of the conditional distribution for the observed history, and $h_b(u)\coloneqq-u\log u-(1-u)\log(1-u)$ is the binary entropy function. All logarithms are natural.
\end{proposition}
This differential entropy bound applies only to conditionally absolutely continuous performance distributions and does not cover point masses arising from invalid LLM responses.
The restriction $\bar\eta_{\mathrm{dir}}<1/2$ ensures that the relaxed entropy bound is
nondecreasing in the miscoverage probability over $[0,\bar\eta_{\mathrm{dir}}]$. The miscoverage probability can thus be replaced by the bound in Eq.~\eqref{eq:uniform_local_eta_definition}. Appendix~\ref{app:proof_local_entropy_bound}
provides the derivation.

For a fixed $\bar\eta_{\mathrm{dir}}$ across the candidate prompts at round $t$, the entropy upper bound in Eq.~\eqref{eq:local_conformal_expected_volume_bound} depends on $p$ only through $h_t^{\mathrm{dir}}(p)$,
defined in Eq.~\eqref{eq:local_conformal_sampling_term}. A smaller penalty therefore yields a smaller entropy upper bound, providing an information theoretic interpretation of the local uncertainty statistic.
\subsection{Conformal-Regularized Excitation Sampling (CERES)}
\label{sec:entropy_regularized_sampling}
The feature excitation score $g_t(p)$ in Eq.~\eqref{eq:gram_excitation_sampling} favors prompts that contribute information in weakly represented feature directions, but does not account for prediction uncertainty. The coverage result in Lemma~\ref{prop:local_knn_coverage} and the entropy bound in Proposition~\ref{prop:local_conformal_entropy_bound} motivate $h_t^{\mathrm{dir}}(p)$ as a local uncertainty penalty after bias compensation. We combine these quantities to address the adaptive prompt sampling Objective~\ref{obj_3} of the problem formulation in Section~\ref{prob:online_llm_performance_learning}.
\subsubsection{CERES Acquisition Rule}
The two scores generally have different numerical scales, so we normalize each over the current candidate set $\mathcal C_t$. The logarithmic uncertainty penalty requires $V_t^{\mathrm{dir}}(p)>0$, which we assume for every $p\in\mathcal C_t$.
Let $\kappa_t(p)$ denote either $g_t(p)$ or $h_t^{\mathrm{dir}}(p)$, and define $\kappa_t^{\mathrm{min}}\coloneqq\min_{p'\in\mathcal C_t}\kappa_t(p')$ and $\kappa_t^{\mathrm{max}}\coloneqq\max_{p'\in\mathcal C_t}\kappa_t(p')$. When $\kappa_t^{\mathrm{max}}>\kappa_t^{\mathrm{min}}$, the normalized score is
\begin{equation}
    \widetilde\kappa_t(p)
    \coloneqq
    \tfrac{
        \kappa_t(p)-\kappa_t^{\mathrm{min}}
    }{
        \kappa_t^{\mathrm{max}}-\kappa_t^{\mathrm{min}}
    }.
    \label{eq:candidate_pool_score_normalization}
\end{equation}
If $\kappa_t^{\mathrm{max}}=\kappa_t^{\mathrm{min}}$, we set $\widetilde\kappa_t(p)=0$ for all $p\in\mathcal C_t$. Since the excitation score depends on $\Sigma_t$ and local uncertainty depends on the current parameter estimate $\widehat\omega_t$ and evaluation memory $\mathcal R_t$, we define the sampling information state used for candidate ranking as $\mathcal S_t\coloneqq(\widehat\omega_t,\Sigma_t,\mathcal R_t)$.

The CERES acquisition score combines normalized feature excitation and local uncertainty:
\begin{equation}
    a_t(p;\mathcal S_t)
    \coloneqq
    \widetilde g_t(p)
    -
    \lambda_h\widetilde h_t^{\mathrm{dir}}(p),
    ~ \lambda_h\geq0,
    \label{eq:value_sampling_acquisition_score}
\end{equation}
and the next prompt is selected according to
\begin{equation}
    p_t
    \in
    \arg\max_{p\in\mathcal C_t}
    a_t(p;\mathcal S_t).
    \label{eq:state_dependent_sampler}
\end{equation}
The choice $\lambda_h=0$ recovers maximum excitation sampling. For $\lambda_h>0$, the second term penalizes candidates with wider local prediction intervals, with $\lambda_h$ controlling the tradeoff between feature excitation and local uncertainty aversion.

\begin{figure}[t]
    \centering
    \includegraphics[width=0.45\textwidth]
        {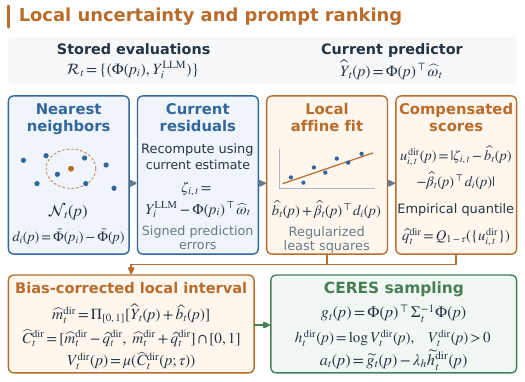}
    \caption{Local bias compensation and uncertainty
    estimation for CERES prompt ranking.}
    \label{fig:ceres_local_sampling}
\end{figure}

Figure~\ref{fig:ceres_local_sampling} visually summarizes the local bias compensation, prediction interval construction, and their combination with feature excitation for prompt selection.
\subsubsection{CERES State Updates}
After CERES selects a prompt $p_t$, the predictor estimates its performance online as $\widehat Y_t=\Phi(p_t)^\top\widehat\omega_t$. Before evaluating the prompt, DtACI constructs a prediction interval around this estimate. Specifically, it maintains $N_{\mathrm e}$ ACI calibration rules with different adaptation gains, called experts. Each expert has a calibration rate $\alpha_t^{(j)}$ and a selection weight $w_t^{(j)}$, collected in $\mathcal A_t\coloneqq\{(\alpha_t^{(j)},w_t^{(j)})\}_{j=1}^{N_{\mathrm e}}$ (see Section~\ref{sec:adaptive_conformal_preliminaries}). DtACI samples one expert according to these weights and uses its rate as $\alpha_t$. This rate determines the quantile from the window $\mathcal W_t^r$, which is used to construct $\widehat C_t(\alpha_t)$ before observing the actual performance $Y_t^{\mathrm{LLM}}$.

After evaluating the prompt, we observe the residual $\xi_t=Y_t^{\mathrm{LLM}}-\widehat Y_t$, its magnitude $s_t=|\xi_t|$, and the calibration feedback $\beta_t$ in Eq.~\eqref{eq:dtaci-beta}. We then update (i)~the predictor parameters via projected squared-loss descent (Eq.~\eqref{eq:online-output-layer-update-squared}); (ii)~the Gram matrix $\Sigma_t$ by adding $\Phi(p_t)\Phi(p_t)^\top$ (Eq.~\eqref{eq:gram_matrix_sampling}); (iii)~the residual window $\mathcal W_t^r$ with $s_t$ for temporal calibration (Eq.~\eqref{eq:global_conformal_interval}) and the evaluation memory $\mathcal R_t$ with the feature--score pair for local bias estimation (Eq.~\eqref{eq:local_directional_fit}); and (iv)~the DtACI expert rates and weights in $\mathcal A_t$ using $\beta_t$ (Section~\ref{sec:adaptive_conformal_preliminaries}). The complete online state for prompt selection and adaptation is thus $Z_t\coloneqq(\widehat\omega_t,\Sigma_t,\mathcal W_t^r,\mathcal R_t,\mathcal A_t)$, with the following updates:
\begin{equation}
\begin{aligned}
    \widehat\omega_{t+1}
    &=
    \Pi_\Omega\left(
        \widehat\omega_t
        +2\eta_t\Phi(p_t)\xi_t
    \right),\\
    \Sigma_{t+1}
    &=
    \Sigma_t+\Phi(p_t)\Phi(p_t)^\top,\\
    \mathcal W_{t+1}^r
    &=
    \operatorname{L
    ast}_r
    \left(
        \mathcal W_t^r\mathbin{\Vert}s_t
    \right),\\
    \mathcal R_{t+1}
    &=
    \mathcal R_t
    \mathbin{\Vert}
    \bigl(\Phi(p_t),Y_t^{\mathrm{LLM}}\bigr),\\
    \mathcal A_{t+1}
    &=
    \operatorname{DtACIUpdate}
    \left(
        \mathcal A_t,\beta_t;
        \Theta_{\mathrm{DtACI}}
    \right).
\end{aligned}
\label{eq:sampling_regret_updates}
\end{equation}
Here, $\operatorname{DtACIUpdate}$ denotes the DtACI expert update for parameters $\Theta_{\mathrm{DtACI}}\coloneqq(\{\gamma_j\}_{j=1}^{N_{\mathrm e}},\eta_{\mathrm{exp}},\sigma)$ with adaptation gains $\gamma_j>0$, weight learning rate $\eta_{\mathrm{exp}}>0$, and weight-sharing parameter $\sigma\in[0,1]$. The operator $\mathbin{\Vert}$ appends an entry to an indexed collection, and $\operatorname{Last}_r$ retains up to the $r$ most recent entries. At the next round, the local residual scores and prediction intervals are recomputed from $\mathcal R_{t+1}$ using the updated parameter estimate $\widehat\omega_{t+1}$, so candidate ranking reflects the current predictor. Table~\ref{tab:online_state_notation} summarizes the notation used in CERES, while Algorithm~\ref{alg:online-sampling-adaptation} presents the complete procedure.

\begin{table}[t]
\centering
\caption{Principal notation for online prediction and CERES sampling.}
\label{tab:online_state_notation}
\footnotesize
\setlength{\tabcolsep}{3pt}
\renewcommand{\arraystretch}{1.05}

\begin{tabularx}{\columnwidth}{
    @{}l>{\raggedright\arraybackslash}X@{}
}
\toprule
Symbol & Meaning \\
\midrule

$\Phi(p),\bar\Phi(p)$
& Augmented prompt features and their normalization \\

$\widehat\omega_t$
& Current output-layer parameter estimate \\

$Y_t^{\mathrm{LLM}},\widehat Y_t$
& Observed and predicted controller performance scores \\

$\xi_t,\ s_t=|\xi_t|$
& Prediction residual at evaluation and its magnitude \\

$\zeta_{i,t}$
& Past observation's residual under the current predictor \\

$\varepsilon_i$
& Noise relative to the conditional mean \\

$\Sigma_t$
& Regularized feature Gram matrix \\

$\mathcal W_t^r$
& Residual window used for temporal calibration \\

$\mathcal R_t$
& Feature and score observations used for local estimation \\

$\mathcal A_t$
& DtACI expert rates and aggregation weights \\

$\mathcal S_t$
& Information used for prompt ranking:
$(\widehat\omega_t,\Sigma_t,\mathcal R_t)$ \\

$Z_t$
& Complete online information:
$(\widehat\omega_t,\Sigma_t,\mathcal W_t^r,
\mathcal R_t,\mathcal A_t)$ \\

$\mathcal N_t(p),K_t$
& Local neighbor index set and its size \\

$g_t(p)$
& Feature excitation score \\

$h_t^{\mathrm{dir}}(p)$
& Logarithm of the local prediction interval width \\

$a_t(p;\mathcal S_t)$
& CERES acquisition score \\

\bottomrule
\end{tabularx}
\end{table}


\begin{algorithm}[htbp]
\caption{CERES with Adaptive Performance Prediction}
\label{alg:online-sampling-adaptation}
\begin{algorithmic}
\renewcommand{\algorithmicrequire}{\textbf{Input:}}
\renewcommand{\algorithmicensure}{\textbf{Output:}}

\Require Feature map $\Phi$, offline parameter
$\omega^{\mathrm{off}}$, feasible set $\Omega$,
prompt sets $\{\mathcal C_t\}_{t=1}^{T}$, and
initial observations $\mathcal D_0$ with $N_0\geq K$;
sampling parameters $\lambda_V,\lambda_h,\lambda_{\mathrm{loc}},
K,r,\{\eta_t\}_{t=1}^{T}$, miscoverage rates
$\alpha,\tau$, and DtACI parameters
$\Theta_{\mathrm{DtACI}}$

\Ensure Online evaluation set $\mathcal Q_T$
and final information state $Z_{T+1}$

\State Set $\widehat\omega_1\leftarrow\omega^{\mathrm{off}}$
and $\mathcal Q_T\leftarrow\emptyset$

\State For $i=1,\ldots,N_0$, compute
$\Phi_i\leftarrow\Phi(p_i)$ and
$s_i\leftarrow
|Y_i^{\mathrm{LLM}}-\Phi_i^\top\widehat\omega_1|$

\State Initialize
$\Sigma_1\leftarrow
\lambda_V I+\sum_{i=1}^{N_0}\Phi_i\Phi_i^\top$

\State Initialize
$\mathcal R_1\leftarrow
\{(\Phi_i,Y_i^{\mathrm{LLM}})\}_{i=1}^{N_0}$
and
$\mathcal W_1^r\leftarrow
\operatorname{Last}_r(\{s_i\}_{i=1}^{N_0})$

\State Initialize
$\mathcal A_1\leftarrow
\bigl((\alpha,1)\bigr)_{j=1}^{N_{\mathrm e}}$

\For{$t=1,\ldots,T$}

    \State Set
    $K_t\leftarrow\min\{K,|\mathcal R_t|\}$
    and
    $\mathcal S_t\leftarrow
    (\widehat\omega_t,\Sigma_t,\mathcal R_t)$

    \State Recompute
    $\zeta_{i,t}\leftarrow
    Y_i^{\mathrm{LLM}}-\Phi_i^\top\widehat\omega_t$
    for all observations in $\mathcal R_t$

    \ForAll{$p\in\mathcal C_t$}

        \State Compute $g_t(p)$ and select
        $\mathcal N_t(p)$ by
        Eqs.~\eqref{eq:gram_excitation_sampling}
        and~\eqref{eq:knn_neighbor_set}

        \State Fit
        $(\widehat b_t(p),
        \widehat{\boldsymbol\beta}_t(p))$
        on $\mathcal N_t(p)$ by
        Eq.~\eqref{eq:local_directional_fit}

        \State Compute the quantile
        $\widehat q_t^{\mathrm{dir}}(p;\tau)$ by
        Eqs.~\eqref{eq:directional_nonconformity}
        and~\eqref{eq:knn_local_nonconformity_quantile}

        \State Construct
        $\widehat C_t^{\mathrm{dir}}(p;\tau)$
        and evaluate $h_t^{\mathrm{dir}}(p)$ by
        Eqs.~\eqref{eq:local_knn_sampling_interval}
        and~\eqref{eq:local_conformal_sampling_term}

    \EndFor

    \State Normalize $g_t$ and $h_t^{\mathrm{dir}}$
    and compute
    $a_t(p;\mathcal S_t)$ by
    Eqs.~\eqref{eq:candidate_pool_score_normalization}
    and~\eqref{eq:value_sampling_acquisition_score}

    \State Select
    $p_t\in\arg\max_{p\in\mathcal C_t}
    a_t(p;\mathcal S_t)$

    \State Draw $I_t$ with $\mathbb P(I_t=j)\propto w_t^{(j)}$ and 
    set $\alpha_t\leftarrow\alpha_t^{(I_t)}$

    \State Compute
    $\widehat Y_t\leftarrow
    \Phi(p_t)^\top\widehat\omega_t$
    and construct $\widehat C_t(\alpha_t)$ using
    Eq.~\eqref{eq:global_conformal_interval}

    \State Evaluate $p_t$ and obtain
    $Y_t^{\mathrm{LLM}}$ by
    Eq.~\eqref{eq:llm_controller_performance}

    \State Compute
    $\xi_t\leftarrow Y_t^{\mathrm{LLM}}-\widehat Y_t$,
    $s_t\leftarrow|\xi_t|$, and
    $\beta_t$ using Eq.~\eqref{eq:dtaci-beta}

    \State Update
    $\widehat\omega_{t+1}$, $\Sigma_{t+1}$,
    $\mathcal W_{t+1}^r$, $\mathcal R_{t+1}$,
    and $\mathcal A_{t+1}$ by
    Eq.~\eqref{eq:sampling_regret_updates}

    \State Append
    $(p_t,\widehat Y_t,\alpha_t,
    \widehat C_t(\alpha_t),Y_t^{\mathrm{LLM}})$
    to $\mathcal Q_T$

\EndFor

\State \Return $\mathcal Q_T$ and $Z_{T+1}$

\end{algorithmic}
\end{algorithm}
\subsection{Finite Sampling Budget Analysis of CERES}
\label{sec:ceres_sampling_guarantee}
We characterize how uncertainty regularization affects the feature excitation achieved by CERES, both at each round and over a finite evaluation budget. We also compare the estimated uncertainty penalty with an oracle reference and examine the excitation guarantees of alternative sampling strategies.
\subsubsection{Per-Round and Finite-Budget Excitation}
Let $p_t^{\mathrm C}\coloneqq p_t$ denote the CERES prompt selection in Eq.~\eqref{eq:state_dependent_sampler}.
For comparison, let $p_t^g\in\arg\max_{p\in\mathcal C_t}g_t(p)$ denote a maximum-excitation prompt, with $m_t\coloneqq g_t(p_t^g)$ being the largest available excitation score. Both selections use the same online state $\mathcal S_t=(\widehat\omega_t,\Sigma_t,\mathcal R_t)$ and candidate prompt set $\mathcal C_t$. Since maximum-excitation sampling corresponds to CERES with $\lambda_h=0$, comparing their scores quantifies the effect of uncertainty regularization. The following theorem bounds the excitation--uncertainty tradeoff at each round and the maximum excitation remaining after $T$ evaluations.
\begin{theorem}[Excitation Uncertainty Tradeoff and Feature Coverage]
\label{thm:ceres_sampling_efficiency}
Suppose that, at each round $t=1,\ldots,T$, $V_t^{\mathrm{dir}}(p)>0$ for every $p\in\mathcal C_t$ and CERES exactly maximizes $\widetilde g_t(p)-\lambda_h\widetilde h_t^{\mathrm{dir}}(p)$ with $0<\lambda_h<1$. The normalized scores are defined by Eq.~\eqref{eq:candidate_pool_score_normalization}. Then the following statements hold.
\begin{enumerate}
    \item \textbf{Excitation uncertainty tradeoff.}
    Define $\Delta g_t \coloneqq \widetilde g_t(p_t^g) - \widetilde g_t(p_t^{\mathrm C})$ and $\Delta h_t \coloneqq \widetilde h_t^{\mathrm{dir}}(p_t^g) - \widetilde h_t^{\mathrm{dir}}(p_t^{\mathrm C})$. Then we get
    \begin{equation}
        0
        \leq
        \Delta g_t
        \leq
        \lambda_h\Delta h_t
        \leq
        \lambda_h.
        \label{eq:ceres_excitation_uncertainty_tradeoff}
    \end{equation}
    Consequently,
    \begin{equation}
    \begin{aligned}
        h_t^{\mathrm{dir}}(p_t^{\mathrm C})
        \leq h_t^{\mathrm{dir}}(p_t^g),
        ~
        g_t(p_t^{\mathrm C})
        \geq (1-\lambda_h)m_t.
    \end{aligned}
    \label{eq:ceres_excitation_uncertainty_consequences}
    \end{equation}
    \item \textbf{Feature coverage under a finite budget.}
    Suppose that $\|\Phi(p)\|_2\leq L_\Phi$ for all
    $p\in\mathcal C_1$. The Gram matrix follows the update in Eq.~\eqref{eq:sampling_regret_updates}, with
    $\Sigma_1\succeq\lambda_V I_{d_\phi+1}$ and $\lambda_V>0$. If $\mathcal C_1$ is finite and $\mathcal C_{t+1} \subseteq \mathcal C_t\setminus\{p_t^{\mathrm C}\}, ~ t=1,\ldots,T$, with $\mathcal C_{T+1}\neq\emptyset$, then, define $G\coloneqq L_\Phi^2/\lambda_V$, a uniform upper bound on excitation. Then
    \begin{equation}
        m_{T+1}
        \leq
        \tfrac{(1+G)(d_\phi+1)}{(1-\lambda_h)T}
        \log\!\left(1+\tfrac{TG}{d_\phi+1}\right).
        \label{eq:ceres_maximum_excitation_rate}
    \end{equation}
\end{enumerate}
\end{theorem}
Eq.~\eqref{eq:ceres_excitation_uncertainty_tradeoff} bounds the loss in normalized excitation by $\lambda_h$. More specifically, Eq.~\eqref{eq:ceres_excitation_uncertainty_consequences} shows that CERES selects a prompt with no greater estimated local uncertainty than the maximum-excitation prompt while retaining at least a fraction $1-\lambda_h$ of its excitation. Also, the finite-budget bound in Eq.~\eqref{eq:ceres_maximum_excitation_rate} controls $m_{T+1}$, the largest excitation score remaining after $T$ evaluations. For a tolerance $\epsilon_g>0$, if the right-hand side is at most $\epsilon_g$, every remaining candidate satisfies $g_{T+1}(p)\leq\epsilon_g$. The factor $(1-\lambda_h)^{-1}$ quantifies the effect of uncertainty regularization on this guarantee. Appendix~\ref{app:proof_ceres_sampling_efficiency} provides the proof.
\subsubsection{Estimated vs. Oracle Uncertainty}
The preceding guarantees use the estimated local uncertainty $h_t^{\mathrm{dir}}(p)$. To quantify its deviation from the uncertainty of the true conditional distribution, we introduce an oracle reference based on the conditional mean $m_\Phi(p)$ and noise quantile $q_{\varepsilon,t}^{\star}(p;\tau)$:
\begin{equation}
    C_{\varepsilon,t}^{\star}(p;\tau)
    \coloneqq
    \left[
        m_\Phi(p)-q_{\varepsilon,t}^{\star}(p;\tau),
        m_\Phi(p)+q_{\varepsilon,t}^{\star}(p;\tau)
    \right]\cap[0,1].
\label{eq:oracle_directional_interval}
\end{equation}
Its width and corresponding uncertainty score are
\begin{equation}
    V_{\varepsilon,t}^{\star}(p)
    \coloneqq
    \mu (C_{\varepsilon,t}^{\star}(p;\tau)),
    ~
    h_{\varepsilon,t}^{\star}(p)
    \coloneqq
    \log V_{\varepsilon,t}^{\star}(p),
\label{eq:oracle_directional_uncertainty}
\end{equation}
where the logarithm requires $V_{\varepsilon,t}^{\star}(p)>0$. Unlike the estimated interval, this reference uses the true conditional mean and noise quantile, allowing us to assess the error in the local uncertainty score.
\begin{corollary}[Oracle Uncertainty Comparison]
\label{cor:ceres_oracle_uncertainty}
Suppose that the conditions for the per-round comparison in Theorem~\ref{thm:ceres_sampling_efficiency}
hold, together with the hypotheses of Lemmas~\ref{lem:directional_residual_separation} and~\ref{lem:local_quantile_error}. Consider a round satisfying $K_t\geq1$ and $\varepsilon_{\mathrm{stat}}(K_t)\leq r_q$, and suppose also that $V_{\varepsilon,t}^{\star}(p)>0$ for every $p\in\mathcal C_t$. We define
$\varepsilon_{h,t}(p) \coloneqq \tfrac{2\Delta_t^{\mathrm{dir}}(p)}{\min\left\{ V_t^{\mathrm{dir}}(p),V_{\varepsilon,t}^{\star}(p)\right\}}$,
where $\Delta_t^{\mathrm{dir}}(p)$ is given in Eq.~\eqref{eq:directional_coverage_perturbation}. On the event $\mathcal E_{\mathrm{conf}}$ from Lemma~\ref{lem:local_quantile_error},
\begin{equation}
    \left|
        h_t^{\mathrm{dir}}(p)
        -
        h_{\varepsilon,t}^{\star}(p)
    \right|
    \leq
    \varepsilon_{h,t}(p)
    \label{eq:directional_uncertainty_approximation}
\end{equation}
for every $p\in\mathcal C_t$. Moreover,
\begin{equation}
    h_{\varepsilon,t}^{\star}(p_t^{\mathrm C})
    \leq
    h_{\varepsilon,t}^{\star}(p_t^g)
    +
    \varepsilon_{h,t}(p_t^{\mathrm C})
    +
    \varepsilon_{h,t}(p_t^g).
\label{eq:ceres_oracle_uncertainty_comparison}
\end{equation}
\end{corollary}
The first bound controls the difference between the estimated and oracle uncertainty scores. The second shows that CERES' preference also holds for the oracle measure, up to estimation errors at the two compared prompts. 
The proof is given in Appendix~\ref{app:proof_ceres_oracle_uncertainty}.
\subsubsection{Comparison with Alternative Sampling Strategies}
\label{rem:ceres_comparison}
We additionally compare CERES with random sampling (RS), K-center diversity sampling (KCDS) based on greedy
core-set selection~\cite{sener2018active}, and local conformal uncertainty sampling (LCUS), using the same online state $\mathcal S_t$ and candidate set $\mathcal C_t$. Theorem~\ref{thm:ceres_sampling_efficiency} already provides the comparison with maximum-excitation sampling.

For RS, $p_t^{\mathrm{RS}}$ is sampled uniformly from $\mathcal C_t$. Let $N_t\coloneqq|\mathcal C_t|$.
Conditioned on $\Sigma_t$ and $\mathcal C_t$, its expected excitation is the average excitation
over the candidate prompts. Since all excitation scores are nonnegative, with one attaining $m_t$, we have
\begin{equation}
    \mathbb E\!\left[
        g_t(p_t^{\mathrm{RS}})
        \,\middle|\,
        \Sigma_t,\mathcal C_t
    \right]
    \coloneqq
    \tfrac{1}{N_t}
    \sum_{p\in\mathcal C_t}g_t(p)
    \geq
    \tfrac{m_t}{N_t}.
\label{eq:random_excitation_comparison}
\end{equation}
This lower bound is approached when one prompt candidate has excitation $m_t$, and the others have
arbitrarily small excitation. Thus, the worst-case expected excitation fraction for RS decreases as $1/N_t$, whereas CERES guarantees a fraction $1-\lambda_h$ independently of $N_t$, as in Eq.~\eqref{eq:ceres_excitation_uncertainty_consequences}.

KCDS instead favors candidates farthest from previous evaluations in its feature space. Let $\mathcal P_{t-1}^{\mathrm{eval}}$ be the initialization prompts and all prompts evaluated before round $t$. It selects
\begin{equation}
    p_t^{\mathrm{KCDS}}
    \in
    \arg\max_{p\in\mathcal C_t}
    \min_{p'\in\mathcal P_{t-1}^{\mathrm{eval}}}
    \left\|
        \bar\Phi(p)-\bar\Phi(p')
    \right\|_2.
    \label{eq:kcds_selection_rule}
\end{equation}
However, geometric separation need not imply high feature excitation. Consider $2n$ distinct evaluated prompts, $n\geq1$, with feature vectors $(1,0,1)^\top$ and $(-1,0,1)^\top$ in equal numbers. With $\lambda_V=1$, these evaluations produce $\Sigma_t=\operatorname{diag}(1+2n,1,1+2n)$.
Let $\mathcal C_t=\{p_a,p_b\}$, with $\Phi(p_a)=(0,0,1)^\top$ and $\Phi(p_b)=(1,1,1)^\top$. Their nearest neighbor distances are $\sqrt{2-\sqrt{2}}$ and $\sqrt{2-4/\sqrt{6}}$, respectively, so KCDS selects $p_a$. The corresponding excitation scores are
$g_t(p_a)=1/(1+2n)$ and $g_t(p_b)=1+2/(1+2n)$.
KCDS therefore achieves an excitation fraction of
$g_t(p_a)/g_t(p_b)=1/(3+2n)$, which approaches zero
as $n\to\infty$.
Hence, KCDS cannot guarantee a uniform positive fraction of maximum excitation under the present assumptions.

LCUS selects prompts solely by local conformal uncertainty. Its original score $h_t^{\mathrm{loc}}(p)$ is the logarithm of the local conformal interval width constructed from residual magnitudes $s_i=|\xi_i|$, giving $p_t^{\mathrm{LCUS}}\in\arg\min_{p\in\mathcal C_t}h_t^{\mathrm{loc}}(p)$. Since this criterion does not account for feature excitation, and the present assumptions impose no relationship between the two scores, the excitation selected by LCUS can be arbitrarily small relative to $m_t$. Thus, a guarantee analogous to Eq.~\eqref{eq:ceres_excitation_uncertainty_consequences} does not follow for LCUS.

In summary, Theorem~\ref{thm:ceres_sampling_efficiency} guarantees that CERES achieves at least a fraction $1-\lambda_h$ of the maximum available excitation at each round. In comparison, RS guarantees a fraction $1/N_t$ in expectation, while the excitation fractions of KCDS and LCUS can approach zero under the present assumptions. Section~\ref{sec:experimental_methodology} compares these methods in terms of held-out prediction accuracy and the number of closed-loop evaluations needed to achieve prescribed accuracy improvements.
\section{Experimental Methodology}
\label{sec:experimental_methodology}
This section presents the experimental setup and methodology for evaluating CERES.
\subsection{Task and Prompt Construction}
\label{sec:prompt_construction}
The experimental task family instantiates Eq.~\eqref{eq:task_dynamics} with control-affine polynomial dynamics. For task $I\in\mathcal I$, the continuous time model $\dot x=f_I(x)+G_I(x)u$ is discretized by forward Euler, $ x_{k+1}=x_k+\Delta t\,[f_I(x_k)+G_I(x_k)u_k]$,
where $x_k\in\mathbb R^n$, $u_k\in\mathbb R^m$, and $\Delta t$ is the sampling interval. The region of interest and admissible input set are $\mathcal X_I=\{x:\|x\|_\infty\leq r\}$ and $\mathcal U_I=\{u:\|u\|_\infty\leq u_{\mathrm{max}}\}$.
\subsubsection{Dynamics Generation}
The offline dataset samples the system dimensions and coefficients, task parameters, and preference profile $\rho$. The polynomial drift and input maps are $f_I(x)\coloneqq \sum_{q=0}^{d_f}\sum_{\gamma\in\mathcal M_q}C_\gamma^{(q)}x^\gamma$ and  $G_I(x)\coloneqq \sum_{q=0}^{d_G}\sum_{\gamma\in\mathcal M_q}D_\gamma^{(q)}x^\gamma$, 
where $\mathcal M_q\coloneqq \{\gamma\in\mathbb N^n:|\gamma|=q\}$, $x^\gamma\coloneqq \prod_{\ell=1}^n x_\ell^{\gamma_\ell}$, $C_\gamma^{(q)}\in\mathbb R^n$, and $D_\gamma^{(q)}\in\mathbb R^{n\times m}$. Coefficients of degree $q$ form arrays $C^{(q)}\in\mathbb R^{M_q\times n}$ and $D^{(q)}\in\mathbb R^{M_q\times n\times m}$, with $M_q=\binom{n+q-1}{q}$.

The sampled polynomial coefficients are conditioned to avoid numerically divergent task instances. Let $A$ denote the linear drift matrix, $A_d=I+\Delta t A$, and let $L_{\mathrm{nl}}$ upper bound the higher-order Jacobian contribution over $\mathcal{B}_r$. The coefficients are rescaled, when necessary, to satisfy $\|A_d\|_2+\Delta t L_{\mathrm{nl}}\leq L_{\mathrm{max}}$. This criterion is used only for numerical task conditioning, not as a global stability certificate. Instances producing nonfinite trajectories or leaving $\mathcal{B}_r$ are discarded.

For each system, we sample $x_0$ and construct $x_{\mathrm{tar}}$ using a bounded nominal control sequence of magnitude $\mu$. The target is $\mathcal X_{T,I}\coloneqq \{x:\|x-x_{\mathrm{tar}}\|_2\leq\epsilon_{\mathrm{goal}}\}$, with level-set function $\phi_{T,I}(x)\coloneqq \|x-x_{\mathrm{tar}}\|_2-\epsilon_{\mathrm{goal}}$. Trajectories over the evaluation horizon $N_I$ are scored using Eq.~\eqref{eq:prelim-performance-score} and Appendix~\ref{app:score}.

In our experiments, the polynomial degrees $d_f$ and $d_G$ are sampled independently and uniformly from $\{2,\ldots,6\}$, and the MPC prediction horizon
$H$ is sampled uniformly from $\{10,\ldots,30\}$. We use $\Delta t=0.1$ and sample $u_{\mathrm{max}}$, $r$, and $\mu$ uniformly from $[0.1,10]$, $[1,10]$, and $[0.15,0.7]$, respectively. The preference weights are sampled independently on a logarithmic scale. The common logarithms of the objective completeness and arrival time weights are uniform on $[-0.3,0.7]$, while those of the control effort and control smoothness weights are uniform on $[-1,0.5]$.
\subsubsection{Task Prompt Design}
\label{subsubsec:prompt-construction}

Each task $I$ is represented by a compact descriptor $\chi(I)=[\chi_{\mathrm{sys}}(I),\chi_{\mathrm{char}}(I)]$, where
\begin{equation}
\begin{aligned}
\chi_{\mathrm{sys}}(I)&\coloneqq[n,m,\Delta t,H,N_I,u_{\mathrm{max}},r,\epsilon_{\mathrm{goal}}],\\
\chi_{\mathrm{char}}(I)&\coloneqq[d_0^{\mathrm{norm}},\rho_{\mathrm{lin}},g_{\mathrm{ctrl}},\eta_{\mathrm{nl}},r_{\mathrm{reach}}].
\end{aligned}
\end{equation}
The characteristic descriptors summarize target distance, linearized stability, input gain, nonlinear dynamics magnitude, and reachability difficulty; Appendix~\ref{app:descriptor-definitions} defines them. The structured prompt $p(I,\rho)=\operatorname{Prompt}(\chi(I),\rho)\in\mathcal S_p$ combines the system instruction, descriptors, initial and target states, preference profile, and JSON specification for seven nonnegative MPC cost weights in the following.
\subsubsection{LLM-Guided MPC}
\label{subsubsec:llm-weight-generation}
The response $a=M_\theta(p)$ and validity indicator follow Section~\ref{sec:llm_controller_evaluator}. A valid response yields $w(p)\coloneqq \operatorname{Parse}_w(a)\in\mathbb R_{\geq0}^7$. Evaluation uses decoding with at most 128 new tokens. The generated cost weight vector is
\begin{equation}
    w
    \coloneqq
    \big[
        W_{\mathrm{track}}^s,\,
        W_{\mathcal X}^s,\,
        W_u^s,\,
        W_{\Delta u}^s,\
        W_{\mathrm{track}}^T,\,
        W_{\mathcal X}^T,\,
        W_u^T
    \big]^\top,
\label{eq:llm-cost-weight-vector}
\end{equation}
where the first four entries weight target tracking, state constraint violation, control effort, and control smoothness. The final three entries weight terminal tracking, terminal state constraint violation, and terminal control effort. The required response JSON format is provided in Appendix~\ref{app:representative-prompt}. Let $x_{\mathrm{tar}}$ denote the target state and let $u_{k-1}$ denote the previously applied input. The stage cost parameterized by $w$ is
\begin{equation}
\begin{aligned}
    \ell_s
    (x_k,u_k,u_{k-1}; w)
    \coloneqq{}&
    W_{\mathrm{track}}^s
    \|x_k-x_{\mathrm{tar}}\|_2^2
    +
    W_{\mathcal X}^s
    \ell_{\mathcal X}(x_k)
    \\
    &+
    W_u^s
    \|u_k\|_2^2
    +
    W_{\Delta u}^s
    \|u_k-u_{k-1}\|_2^2,
\end{aligned}
\label{eq:stage-cost}
\end{equation}
where $\ell_{\mathcal X}(x) \coloneqq \sum_{q=1}^{n} \left[ \max \left\{ 0,\,|x_q|-r \right\} \right]^2$ is a soft penalty for leaving the region of interest $\mathcal B_r$. This penalty is zero for $x\in\mathcal B_r$ and increases quadratically with the magnitude of the coordinate-wise constraint violations. The terminal cost is
\begin{equation}
\begin{aligned}
    \ell_T
    (x_H,u_{H-1}; w)
    \coloneqq{}&
    W_{\mathrm{track}}^T
    \|x_H-x_{\mathrm{tar}}\|_2^2
    \\
    &+
    W_{\mathcal X}^T
    \ell_{\mathcal X}(x_H)
    +
    W_u^T
    \|u_{H-1}\|_2^2.
\end{aligned}
\label{eq:terminal-cost}
\end{equation}
At control step $k$, the generated weights parameterize the following finite-horizon MPC problem, with $\bar u_{-1}=u_{k-1}$:
\begin{align}
&\oldmin_{\bar{x}_{0:H},\,\bar{u}_{0:H-1}}
\sum_{\ell=0}^{H-1}
\ell_s(\bar{x}_\ell,\bar{u}_\ell,\bar{u}_{\ell-1};w)
+
\ell_T(\bar{x}_H,\bar{u}_{H-1};w)\nonumber
\\
&\mathrm{s.t.}~
\bar{x}_0=x_k,~
\bar{u}_{\ell}\in\mathcal U,
~\ell=0,\ldots,H-1,
\\
&\textcolor{white}{\mathrm{s.t.}~}
\bar{x}_{\ell+1}
=
\bar{x}_{\ell}
+
\Delta t
\left[
f_I(\bar{x}_{\ell})
+
G_I(\bar{x}_{\ell})\bar{u}_{\ell}
\right].
\label{eq:llm_mpc_problem}
\end{align}
Let $\{\bar{x}^{\star}_{0:H},\bar{u}^{\star}_{0:H-1}\}$ denote a locally optimal solution of Eq.~\eqref{eq:llm_mpc_problem}. Only the first optimized input, $u_k=\bar{u}^{\star}_0$, is applied to the system, and the optimization
is repeated at the next measured state. Over the closed loop evaluation
horizon $N_I$, this procedure produces $X^p=\{x_k\}_{k=0}^{N_I}$ and $U^p=\{u_k\}_{k=0}^{N_I-1}$, which are evaluated using Eq.~\eqref{eq:llm_controller_performance}.
\subsubsection{Offline Cost Weight Search via iLQR-MPC}
\label{subsubsec:offline-weight-search}
We solve Eq.~\eqref{eq:llm_mpc_problem} using standard iLQR~\cite{li2004iterative,6386025}. To include the input and smoothness term, the state is augmented as $z_k=[x_k^\top,u_{k-1}^\top]^\top\in\mathbb R^{n+m}$, with dynamics
\begin{equation}
    z_{k+1}=\widetilde F_I(z_k,u_k)
    =\begin{bsmallmatrix}
       x_k+\Delta t[f_I(x_k)+G_I(x_k)u_k]\\ u_k
     \end{bsmallmatrix}.
\end{equation}
Each MPC problem is solved with at most five iLQR iterations, using a regularized backward pass and line search over forward rollouts with inputs projected onto the admissible set. For each task and preference pair $(I_i,\rho_i)$, we construct $p_i\coloneqq p(I_i,\rho_i)$ and sample $N_w$ candidate weight vectors $\mathcal W_i\coloneqq \{w_i^{(j)}\}_{j=1}^{N_w}\subset\mathbb R_{\geq0}^7$. Closed-loop iLQR-MPC evaluation produces $(X_i^{(j)},U_i^{(j)})=\mathcal E_{\mathrm{iLQR}}(I_i,w_i^{(j)})$ and score $Y_i^{(j)}\coloneqq Y(X_i^{(j)},U_i^{(j)};I_i,\rho_i)$. We select $w_i^\star\coloneqq w_i^{(j_i^\star)}$, where $j_i^\star\in\arg\max_{1\leq j\leq N_w}Y_i^{(j)}$.
Thus $w_i^\star$ is the best weight within the sampled set. For optional fine-tuning, we could construct $\mathcal D_{\mathrm{SFT}}\coloneqq\{(p_i,w_i^\star,Y_i^\star)\}_{i=1}^{N_{\mathrm{SFT}}}$, where $w_i^\star$ is the training target for $p_i$, and $Y_i^\star=Y_i^{(j_i^\star)}$ records its evaluated performance.
\subsection{Performance Prediction for LLM-Guided Controllers}
\label{sec:performance_predictor_training}
Evaluating an LLM-generated MPC weight vector requires a complete closed-loop rollout. We therefore train a predictor that maps the structured prompt directly to the resulting score $Y_i^{\mathrm{LLM}}$ defined in Section~\ref{sec:llm_controller_evaluator}.
\subsubsection{Optional LLM Fine-Tuning}
\label{sec:offline_llm_training}
Although CERES does not require LLM fine-tuning, we may consider a two-stage adaptation of \emph{Llama-3.2-1B-Instruct}~\cite{grattafiori2024llama}. Supervised fine-tuning (SFT) can first learn the mapping from nominally available prompts to selected MPC cost weights in the prescribed JSON format. Group Relative Policy Optimization (GRPO) then refines the SFT-initialized LLM using rewards based on the closed-loop performance of newly generated MPC weights.

For each entry of $\mathcal D_{\mathrm{SFT}}$, the target response $a_i^\star$ represents the selected MPC weights $w_i^\star$ as the JSON object $\{\texttt{"w"}:w_i^\star\}$. Its token sequence is $(a_{i,1}^\star,\ldots,a_{i,L_i}^\star)$, where $a_{i,\ell}^\star$ denotes the $\ell$th token and $L_i$ is the number of response tokens.
SFT minimizes the response token cross entropy by
\begin{equation}
\begin{aligned}
    \mathcal{L}_{\mathrm{SFT}}(\theta)
    \coloneqq-
    \tfrac{1}{N_{\mathrm{SFT}}}
    \sum_{i=1}^{N_{\mathrm{SFT}}}
    \tfrac{1}{L_i}
    \sum_{\ell=1}^{L_i}
    \log
    P_{\theta}
    \left(
        a_{i,\ell}^\star
        \mid
        p_i,a_{i,<\ell}^\star
    \right),
\end{aligned}
\label{eq:sft-objective}
\end{equation}
where $a_{i,<\ell}^\star$ denotes the preceding target tokens. GRPO then optimizes the performance of newly generated weights using $\mathcal D_{\mathrm{GRPO}}=\{(I_i,\rho_i,p_i)\}_{i=1}^{N_{\mathrm{GRPO}}}$,
where $I_i$ is the control task instance,
$\rho_i$ specifies the performance priorities,
$p_i$ is the corresponding structured prompt,
and $N_{\mathrm{GRPO}}$ is the number of training instances. For each prompt, it samples $a_i^{(g)}\sim P_{\theta_{\mathrm{old}}}(\cdot\mid p_i)$, $g=1,\ldots,N_g$, where $\theta_{\mathrm{old}}$ contains the generation-policy parameters. Responses are assigned validity indicators $\nu_i^{(g)}$ and scores $Y_i^{(g)}$ by Eq.~\eqref{eq:llm_controller_performance}: valid weights undergo closed-loop iLQR-MPC evaluation, and invalid responses receive zero.
The GRPO reward $R_i^{(g)}$ combines response validity and closed loop performance is 
\begin{equation}
    R_i^{(g)}
    \coloneqq
    r_{\mathrm{valid}}\nu_i^{(g)}
    -r_{\mathrm{invalid}}(1-\nu_i^{(g)})
    +\lambda_Y Y_i^{(g)},
    \label{eq:grpo-reward}
\end{equation}
where $\lambda_Y,r_{\mathrm{valid}},r_{\mathrm{invalid}}\geq0$. 
We otherwise use the standard clipped GRPO objective~\cite{shao2024deepseekmath}, with group-relative advantages computed from Eq.~\eqref{eq:grpo-reward} and KL regularization toward the response distribution of the SFT-initialized LLM.
\subsubsection{Performance Predictor for Online Adaptation}
\begin{figure}[htbp]
    \centering
    \includegraphics[width=0.45\textwidth]{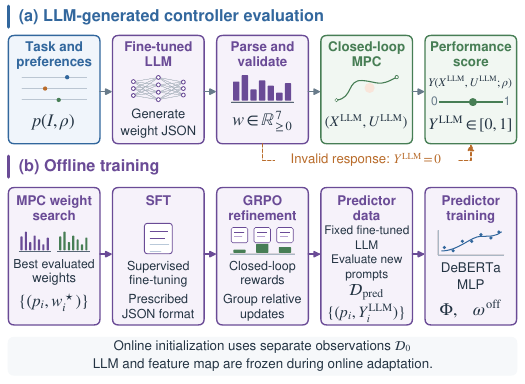}
    \caption{Controller evaluation and LLM \& predictor training.}
    \label{fig:ceres_controller_training}
\end{figure}

CERES uses a fixed prompt feature map with an output layer that adapts online, as introduced in Section~\ref{sec:online_prediction_preliminaries}. For the optional predictor initialization used in our experiments, we use $\mathcal D_{\mathrm{pred}}\coloneqq\{(p_i,Y_i^{\mathrm{LLM}})\}_{i=1}^{N_{\mathrm{pred}}}$, consisting of previously evaluated prompt--performance pairs. Figure~\ref{fig:ceres_controller_training} illustrates the evaluation pipeline, including optional LLM refinement and predictor initialization.

We instantiate the prompt feature map using \emph{deberta-v3-base} with a compact score prediction head. Inputs are truncated to 1,024 tokens. For contextual token embeddings $h_{i,\ell}\in\mathbb R^d$, masked mean pooling followed by a hidden transformation gives
\begin{equation}
    e_i\coloneqq\tfrac{\sum_{\ell=1}^{L_i}m_{i,\ell}h_{i,\ell}}
                   {\sum_{\ell=1}^{L_i}m_{i,\ell}},
    ~
    \varphi_i=\operatorname{GELU}(W_he_i+b_h),
\end{equation}
where $m_{i,\ell}\in\{0,1\}$ excludes padding, $W_h\in\mathbb R^{d_\phi\times d}$, and $b_h\in\mathbb R^{d_\phi}$. The encoder and hidden transformation define $\varphi_i=\phi_{\bar\psi}(p_i)\in\mathbb R^{d_\phi}$. With output weights $v_o\in\mathbb R^{d_\phi}$ and bias $b_o\in\mathbb R$, we set $\Phi_{\bar\psi}(p_i)=[\varphi_i^\top,1]^\top$ and $\omega=[v_o^\top,b_o]^\top$, yielding $\widehat Y_{\bar\psi,\omega}(p_i)=\Phi_{\bar\psi}(p_i)^\top\omega$, consistent with Section~\ref{sec:online_prediction_preliminaries}.

For this optional initialization, we could jointly fit the encoder and score head using the regularized smooth-L1 objective
\begin{equation}
\begin{aligned}
    \mathcal L_{\mathrm{pred}}(\bar\psi,\omega)
    \coloneqq{}&\tfrac{1}{N_{\mathrm{pred}}}\sum_{i=1}^{N_{\mathrm{pred}}}
    \ell_{\mathrm{SL1}} (
    \widehat Y_{\bar\psi,\omega}(p_i),Y_i^{\mathrm{LLM}} )\\
    &+\lambda_{\mathrm{score}}\|\omega\|_2^2,
\end{aligned}
\label{eq:predictor-loss}
\end{equation}
where $\lambda_{\mathrm{score}}\geq0$ and, for $e=\widehat Y-Y$,
\begin{equation}
    \ell_{\mathrm{SL1}}(\widehat Y,Y)\coloneqq
    \begin{cases}
        e^2/(2\delta_{\mathrm{SL1}}),&|e|<\delta_{\mathrm{SL1}},\\
        |e|-\delta_{\mathrm{SL1}}/2,&|e|\geq\delta_{\mathrm{SL1}},
    \end{cases}
    \label{eq:predictor-smooth-l1-loss}
\end{equation}
with transition parameter $\delta_{\mathrm{SL1}}>0$. Dropout is applied around the hidden transformation during training. The resulting parameters $(\bar\psi^\star,\omega^{\mathrm{off}})$ provide the fixed feature map $\Phi=\Phi_{\bar\psi^\star}$ and initial estimate $\widehat\omega_1=\omega^{\mathrm{off}}$; subsequent CERES adaptation updates only the output layer.
\subsection{Heterogeneous Robot Hardware Evaluation}
\label{sec:hardware_implementation}
\begin{figure}[t]
    \centering
    \includegraphics[width=0.8\columnwidth]
        {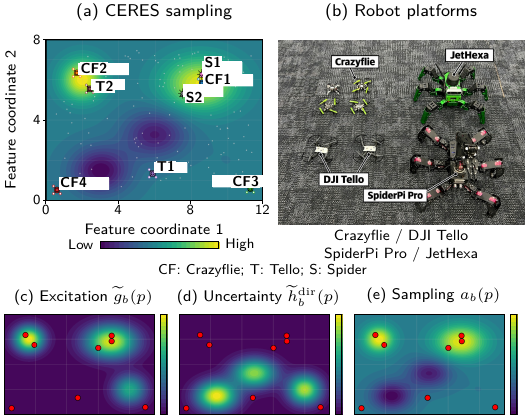}
    \caption{CERES sampling and robot platforms for multi-robot evaluation.}
    \label{fig:multi_robot_setup}
\end{figure}

The hardware experiments use multiple aerial robots (\myhref{https://www.bitcraze.io/products/crazyflie-2-1-brushless/}{Crazyflies} and \myhref{https://www.ryzerobotics.com/tello}{DJI Tellos}) and legged robots (\myhref{http://wiki.hiwonder.com/projects/SpiderPi_Pro/en/latest/docs/1_read_before_studying.html}{SpiderPi Pro} and \myhref{https://www.hiwonder.com/products/jethexa-2027?variant=44274869731415}{JetHexa}), as shown in Fig.~\ref{fig:multi_robot_setup}(b). Figure~\ref{fig:multi_robot_setup}(a) illustrates example prompt assignments in schematic feature coordinates. Figure~\ref{fig:multi_robot_setup}(c)--(e) show the corresponding feature excitation, local uncertainty, and sampling scores used by CERES.
A \myhref{https://www.qualisys.com/}{Qualisys} motion-capture system provides a common external pose reference for trajectory tracking and closed-loop score evaluation. An external computer performs online prompt selection, LLM inference, MPC trajectory planning, and batch coordination, while each platform uses its native low-level stabilization or gait controller. The platform class set is defined as $\mathcal{K}\coloneqq \{\text{Crazyflie},\text{DJI Tello}, \text{SpiderPi},\text{JetHexa}\}$.

For hardware evaluation, let $X^{\mathrm{hw}}\coloneqq(x_k^{\mathrm{hw}})_{k=0}^{N_I}$ denote the task state trajectory reconstructed from Qualisys measurements, and let $U^{\mathrm{hw}}\coloneqq(u_k^{\mathrm{hw}})_{k=0}^{N_I-1}$ denote the recorded command sequence transmitted to the assigned robot. For a valid LLM response, the hardware performance score is $Y^{\mathrm{LLM}}(p) =Y(X^{\mathrm{hw}},U^{\mathrm{hw}};I,\rho)$.
\subsubsection{Platform-Specific Control Interfaces}
The planned trajectories are executed differently across platforms, depending on their onboard control capabilities.

\noindent\emph{\underline{Crazyflies}}.
Each Crazyflie receives Qualisys pose measurements through the external computer and uses them as external state-estimator updates. The computer supplies position/attitude references, while the onboard firmware performs state estimation, attitude stabilization, and motor control.

\noindent\emph{\underline{DJI Tellos}}.
The Tello accepts high-level body-frame SDK commands rather than the full MPC trajectory. We therefore use Qualisys feedback in an external tracking loop that converts the planned trajectory into body-frame RC commands.

\noindent\emph{\underline{Legged Robots}}.
SpiderPi Pro and JetHexa receive high-level planar locomotion commands computed from the reference trajectory and Qualisys feedback. SpiderPi uses its native gait interface and JetHexa its ROS interface; their onboard controllers perform gait generation and actuator-level control.
\subsubsection{Heterogeneous Multi-Robot Evaluation}
\label{sec:multi_robot}

We extend the single query procedure to a heterogeneous multi-robot setting in which multiple prompts are evaluated simultaneously. At batch $b$, $M_b$ is the number of available robots, and $\mathcal M_b\coloneqq\{1,\ldots,M_b\}$ is their index set. Let $\mathcal K$ denote the set of platform
classes. The metadata map $\chi_{\mathrm{plat}}(p)\in\mathcal K$ specifies the platform class required by prompt $p$, while $\chi_{\mathrm{rob},b}(m)\in\mathcal K$ identifies the platform class of robot $m$.
Multiple robots may share a platform class.
Robot $m$ can evaluate prompt $p$ when
$\chi_{\mathrm{plat}}(p)=\chi_{\mathrm{rob},b}(m)$.
For class $k$, let $c_{b,k}\coloneqq|\{m\in\mathcal M_b: \chi_{\mathrm{rob},b}(m)=k\}|$ denote the available robot capacity. Batch index $b$ replaces $t$ in the online state $Z_t$ and ranking state $\mathcal S_t$ defined in Section~\ref{sec:entropy_regularized_sampling}.
\subsubsection{Compatibility Constrained Batch Selection}
\label{sec:compatibility_constrained_batch_selection}
At each~$b$, prediction residuals are recomputed from $\mathcal R_b$ using $\widehat\omega_b$. The local uncertainty statistic is then evaluated for each candidate $p\in\mathcal C_b$ following Section~\ref{sec:local_conformal_analysis}. The acquisition scores $a_b(p;\mathcal S_b)$ are computed using
Eq.~\eqref{eq:value_sampling_acquisition_score} and remain fixed throughout batch sampling.

Candidates are processed in decreasing order of $a_b(p;\mathcal S_b)$ and accepted whenever their platform class has remaining capacity. The selected batch $\mathcal B_b$ therefore satisfies
\begin{equation}
    |\mathcal B_b|\le M_b,~
    \sum_{p\in\mathcal B_b}
    \mathbf 1\{\chi_{\mathrm{plat}}(p)=k\}
    \le c_{b,k},~ k\in\mathcal K.
\label{eq:heterogeneous_batch_selection}
\end{equation}
Selection terminates when all available robots are assigned or no compatible candidate remains. Capacity feasibility guarantees a compatible one to one assignment of the selected prompts to available robots. We denote the resulting robot prompt pairs by $\mathcal P_b\coloneqq\{(m_{b,j},p_{b,j})\}_{j=1}^{n_b}$, where
$n_b=|\mathcal B_b|\leq M_b$ and  $\chi_{\mathrm{rob},b}(m_{b,j})
=\chi_{\mathrm{plat}}(p_{b,j})$
\subsubsection{Parallel Execution and Sequential Calibration}
\label{sec:multi_robot_execution_calibration}
The assignments in $\mathcal P_b$ are executed in parallel. The predictor is fixed within the batch, giving $\widehat Y_{b,j} = \Phi(p_{b,j})^\top\widehat\omega_b$.
After execution, the observations are processed in a predetermined order $j=1,\ldots,n_b$, with $\xi_{b,j} = Y_{b,j}^{\mathrm{LLM}} - \widehat Y_{b,j}$ and $s_{b,j} = |\xi_{b,j}|$.
Starting from $(\mathcal A_{b,1},\mathcal W_{b,1}^{r}) =(\mathcal A_b,\mathcal W_b^{r})$, the observations are processed in the predetermined order. For observation $j$, the calibration rate $\alpha_{b,j}$ is selected from $\mathcal A_{b,j}$, and the prediction interval is constructed using $\mathcal W_{b,j}^{r}$ and the fixed prediction $\widehat Y_{b,j}$. The score $Y_{b,j}^{\mathrm{LLM}}$ is then incorporated to update the expert state and residual window using the DtACI procedure in Algorithm~\ref{alg:online-sampling-adaptation}. 
The final expert state and residual window initialize the next batch.
\subsubsection{Batch Predictor and Sampling-State Update}
\label{sec:multi_robot_batch_state_update}
After processing all $n_b$ observations, we update the output parameter estimate, feature Gram matrix, and
evaluation memory. The parameter update is a projected gradient step for the mean squared prediction loss over
the completed batch. For $n_b\geq1$, the updates are
\begin{equation}
\begin{aligned}
    \widehat\omega_{b+1}
    &=
    \Pi_\Omega
    \left[
        \widehat\omega_b
        +
        \tfrac{2\eta_b}{n_b}
        \sum_{j=1}^{n_b}
        \Phi(p_{b,j})\xi_{b,j}
    \right],
    \\
    \Sigma_{b+1}
    &=
    \Sigma_b
    +
    \sum_{j=1}^{n_b}
    \Phi(p_{b,j})\Phi(p_{b,j})^\top,
    \\
    \mathcal R_{b+1}
    &=
    \mathcal R_b
    \mathbin{\Vert}
    \left(
        \bigl(\Phi(p_{b,j}),Y_{b,j}^{\mathrm{LLM}}\bigr)
    \right)_{j=1}^{n_b}.
\end{aligned}
\label{eq:heterogeneous_batch_online_update}
\end{equation}
If no feasible candidate is available, the batch round is skipped, and the online variables remain unchanged. Initialization, scoring, and DtACI follow Algorithm~\ref{alg:online-sampling-adaptation}, while the multi-robot procedure selects compatible prompts for parallel evaluation and updates the predictor after each batch.
\subsection{Evaluation Protocol for Online Sampling Strategies}
\label{sec:online_evaluation_metric}
Single-query experiments follow Algorithm~\ref{alg:online-sampling-adaptation}, while heterogeneous multi-robot experiments follow Section~\ref{sec:multi_robot}. 
DtACI uses target miscoverage rate $\alpha=0.2$, residual window $r=25$, and $N_e=7$ expert gains logarithmically spaced over $[10^{-3},5\times10^{-2}]$, with weight parameters $\eta_{\mathrm{exp}}=0.01195$ and $\sigma=10^{-3}$. Local uncertainty estimation uses $K=200$ nearest neighbors, a local miscoverage rate $\tau=0.3$, and a regression regularization parameter $\lambda_{\mathrm{loc}}=10^{-3}$. The feature Gram matrix is initialized with $\lambda_V=1$. 

We then compare CERES with RS, KCDS, and LCUS from Remark~\ref{rem:ceres_comparison}, and evaluate CERES for $\lambda_h\in\{0,0.1,0.5\}$, where $\lambda_h=0$ recovers excitation-only sampling. All methods share the same initial predictor, initialization dataset, candidate prompt pool, online update rules, DtACI settings, and evaluation budgets. Multi-robot experiments also use identical robot availability and compatibility constraints, so only the prompt selection rule differs across methods.

Prediction accuracy is evaluated on the fixed held-out set using mean squared error (MSE) and the coefficient of determination ($R^2$), with relative improvements measured against the initial predictor. Coverage is the fraction of held-out scores contained in their corresponding prediction intervals, with target coverage $1-\alpha=0.8$. At each checkpoint, the predictor is frozen, and a separate calibration state is initialized from the current DtACI state and residual window. Held-out observations are processed sequentially, constructing each interval before using its corresponding score to update this evaluation state. Sample efficiency is measured by the earliest budget required to achieve a prescribed MSE reduction or $R^2$ increase, counting individual controller evaluations for single-robot experiments and batches of compatible evaluations for multi-robot experiments.
\section{Results and Discussion}
\label{sec:result}
We assess how effectively CERES improves performance prediction under limited controller evaluation budgets. We first examine sequential single-robot adaptation, where each observation requires one closed-loop evaluation, followed by heterogeneous multi-robot experiments involving platform compatibility and parallel execution. We also examine how uncertainty regularization affects sampling decisions and how estimated uncertainty relates to observed prediction errors. Table~\ref{tab:representative_prompts} presents
representative task prompts used in the experiments,
including their control objectives and performance priorities.

\begin{table}[htbp]
\centering
\caption{Illustrative structured task prompt}
\label{tab:representative_prompts}
\footnotesize
\setlength{\tabcolsep}{3pt}
\renewcommand{\arraystretch}{1.08}

\begin{tabularx}{\columnwidth}{
    @{}>{\raggedright\arraybackslash}p{0.28\columnwidth}
    >{\raggedright\arraybackslash}X@{}
}
\toprule
Component & Prompt content \\
\midrule

Instruction
&
Generate MPC cost weights for the specified
target-reaching task and performance priorities.
\\
\addlinespace[2pt]

Task dimensions
&
State dimension $n=3$; input dimension $m=1$.
\\
\addlinespace[2pt]

Time parameters
&
Time step $\Delta t=0.05$; MPC horizon $H=10$;
evaluation horizon $N_I=40$.
\\
\addlinespace[2pt]

Bounds and target
&
Input bound $u_{\max}=2$; region radius $r=5$;
target tolerance $\epsilon_{\mathrm{goal}}=0.05$.
\\
\addlinespace[2pt]

Performance priorities
&
Objective completeness: $2$; arrival time: $1.5$;
control effort: $0.5$; control smoothness: $0.3$.
\\
\addlinespace[2pt]

Required response
&
A single JSON object:
\texttt{\{"w":[}$w_1,\ldots,w_7$\texttt{]\}},
with seven finite nonnegative MPC cost weights.
\\

\bottomrule
\end{tabularx}
\end{table}

The initial predictor shared across all methods achieves a held-out MSE of $0.0642$ and $R^2=0.1255$. In the tables, bold entries indicate the lowest MSE, highest $R^2$, or fewest evaluations or rounds, including ties at the displayed precision. N/R indicates that an improvement target was not reached within the reported evaluation budget.
\subsection{Single Robot Prediction Performance and Efficiency}
For objectives~\ref{obj_1} and~\ref{obj_3}, we assess held-out prediction accuracy and evaluation efficiency, complementing the regret analysis in Theorem~\ref{thm:online-ls-regret}. 
At least one CERES configuration attains the lowest MSE and highest $R^2$ at every budget in Table~\ref{tab:single_robot_fixed_budget}. At $50$ evaluations, all three tie at an MSE of $0.0582$ to the displayed precision; $\lambda_h=0.1$ gives the highest $R^2$ ($0.2070$). At $100$ evaluations, $\lambda_h=0.5$ leads both metrics (MSE $0.0579$, $R^2=0.2111$). At $400$, $\lambda_h=0.1$ gives an MSE of $0.0578$ and $R^2=0.2130$, $10.0\%$ lower MSE and $69.7\%$ higher $R^2$ than the offline predictor.
\begin{table}[htbp]
\centering
\caption{Single-robot prediction performance at fixed
evaluation budgets.
$C_{\lambda}$ denotes CERES with $\lambda_h=\lambda$.}
\label{tab:single_robot_fixed_budget}

\scriptsize
\setlength{\tabcolsep}{0.5pt}
\renewcommand{\arraystretch}{0.94}

\begin{minipage}[t]{0.495\columnwidth}
\vspace{0pt}
\centering
\begin{tabular*}{\linewidth}{
@{\extracolsep{\fill}}c l c c c@{}
}
\toprule
Eval. & Method & MSE & $R^2$ & Cov. \\
\midrule
0 & Offline & 0.0642 & 0.1255 & -- \\
\midrule

\multirow{6}{*}{50}
& RS        & 0.0590 & 0.1966 & 0.7875 \\
& KCDS      & 0.0584 & 0.2041 & 0.7900 \\
& LCUS      & 0.0587 & 0.1997 & 0.7900 \\
& $C_0$     & \textbf{0.0582} & 0.2069 & 0.7925 \\
& $C_{0.1}$ & \textbf{0.0582} & \textbf{0.2070} & 0.7900 \\
& $C_{0.5}$ & \textbf{0.0582} & 0.2067 & 0.7900 \\
\addlinespace[2pt]

\multirow{6}{*}{100}
& RS        & 0.0595 & 0.1893 & 0.7950 \\
& KCDS      & 0.0580 & 0.2092 & 0.7875 \\
& LCUS      & 0.0586 & 0.2015 & 0.7975 \\
& $C_0$     & 0.0581 & 0.2090 & 0.7925 \\
& $C_{0.1}$ & 0.0580 & 0.2101 & 0.7975 \\
& $C_{0.5}$ & \textbf{0.0579} & \textbf{0.2111} & 0.8050 \\
\addlinespace[2pt]

\multirow{6}{*}{150}
& RS        & 0.0594 & 0.1901 & 0.7900 \\
& KCDS      & 0.0582 & 0.2073 & 0.7950 \\
& LCUS      & 0.0586 & 0.2021 & 0.7925 \\
& $C_0$     & \textbf{0.0579} & 0.2105 & 0.7950 \\
& $C_{0.1}$ & \textbf{0.0579} & \textbf{0.2115} & 0.8025 \\
& $C_{0.5}$ & \textbf{0.0579} & 0.2109 & 0.8075 \\
\bottomrule
\end{tabular*}
\end{minipage}%
\hfill%
\begin{minipage}[t]{0.495\columnwidth}
\vspace{0pt}
\centering
\begin{tabular*}{\linewidth}{
@{\extracolsep{\fill}}c l c c c@{}
}
\toprule
Eval. & Method & MSE & $R^2$ & Cov. \\
\midrule
0 & Offline & 0.0642 & 0.1255 & -- \\
\midrule

\multirow{6}{*}{200}
& RS        & 0.0594 & 0.1909 & 0.8000 \\
& KCDS      & 0.0584 & 0.2043 & 0.7900 \\
& LCUS      & 0.0593 & 0.1916 & 0.7900 \\
& $C_0$     & 0.0590 & 0.1956 & 0.7925 \\
& $C_{0.1}$ & \textbf{0.0580} & \textbf{0.2104} & 0.7950 \\
& $C_{0.5}$ & 0.0581 & 0.2083 & 0.7900 \\
\addlinespace[2pt]

\multirow{6}{*}{300}
& RS        & 0.0593 & 0.1921 & 0.8000 \\
& KCDS      & 0.0582 & 0.2077 & 0.8000 \\
& LCUS      & 0.0592 & 0.1933 & 0.8000 \\
& $C_0$     & 0.0581 & 0.2080 & 0.7975 \\
& $C_{0.1}$ & 0.0580 & 0.2103 & 0.8000 \\
& $C_{0.5}$ & \textbf{0.0579} & \textbf{0.2110} & 0.7950 \\
\addlinespace[2pt]

\multirow{6}{*}{400}
& RS        & 0.0591 & 0.1954 & 0.7900 \\
& KCDS      & 0.0585 & 0.2036 & 0.8050 \\
& LCUS      & 0.0592 & 0.1932 & 0.8050 \\
& $C_0$     & 0.0580 & 0.2093 & 0.7975 \\
& $C_{0.1}$ & \textbf{0.0578} & \textbf{0.2130} & 0.7925 \\
& $C_{0.5}$ & 0.0581 & 0.2085 & 0.7975 \\
\bottomrule
\end{tabular*}
\end{minipage}
\end{table}

Table~\ref{tab:sample_efficiency}(a) presents the number of individual controller evaluations required to first attain a prescribed prediction improvement. All three CERES configurations reach a $9\%$ MSE reduction after $44$ evaluations, compared with $51$ for KCDS and $126$ for LCUS. This represents $13.7\%$ fewer
evaluations than KCDS and $65.1\%$ fewer than LCUS. RS does not reach this target within the reported evaluations. A $60\%$ relative increase in $R^2$ is achieved after $39$ evaluations by CERES, $43$ by KCDS, and $55$ by LCUS, while RS again does not reach the target. CERES therefore achieves the best evaluation efficiency among the compared sampling strategies for both prediction improvement targets, requiring the fewest closed-loop controller evaluations.

The identical threshold counts for the three CERES configurations show the substantial contribution of feature excitation during initial adaptation. The effect of uncertainty regularization becomes visible at later budgets. At $200$ evaluations, for example, $\lambda_h=0.1$ gives an MSE of $0.0580$ and $R^2=0.2104$, whereas excitation only acquisition gives $0.0590$ and $0.1956$. This comparison demonstrates the additional value of local residual
information when selecting among prompts that contribute feature excitation.

\begin{table}[htbp]
\centering
\caption{Evaluations and rounds required to attain prescribed
prediction improvements.
$C_{\lambda}$ denotes CERES with $\lambda_h=\lambda$.}
\label{tab:sample_efficiency}

\scriptsize
\setlength{\tabcolsep}{0.5pt}
\renewcommand{\arraystretch}{1.0}

\begin{minipage}[t]{0.49\columnwidth}
\centering
(a) Single-robot evaluations
\par\smallskip

\begin{tabular*}{\linewidth}{
@{\extracolsep{\fill}}lcccccc@{}
}
\toprule
& \multicolumn{3}{c}{MSE reduction}
& \multicolumn{3}{c}{$R^2$ increase} \\
\cmidrule(lr){2-4}
\cmidrule(lr){5-7}
Method
& $7\%$ & $8\%$ & $9\%$
& $40\%$ & $50\%$ & $60\%$ \\
\midrule

RS
& 32 & 41 & N/R
& 28 & 33 & N/R \\

KCDS
& 32 & 38 & 51
& \textbf{27} & 33 & 43 \\

LCUS
& 32 & 41 & 126
& \textbf{27} & 33 & 55 \\

$C_0$
& \textbf{30} & \textbf{35} & \textbf{44}
& \textbf{27} & \textbf{31} & \textbf{39} \\

$C_{0.1}$
& \textbf{30} & \textbf{35} & \textbf{44}
& \textbf{27} & \textbf{31} & \textbf{39} \\

$C_{0.5}$
& \textbf{30} & \textbf{35} & \textbf{44}
& \textbf{27} & \textbf{31} & \textbf{39} \\

\bottomrule
\end{tabular*}
\end{minipage}%
\hfill%
\begin{minipage}[t]{0.49\columnwidth}
\centering
(b) Multi-robot rounds
\par\smallskip

\begin{tabular*}{\linewidth}{
@{\extracolsep{\fill}}lcccccc@{}
}
\toprule
& \multicolumn{3}{c}{MSE reduction}
& \multicolumn{3}{c}{$R^2$ increase} \\
\cmidrule(lr){2-4}
\cmidrule(lr){5-7}
Method
& $5\%$ & $7\%$ & $9\%$
& $50\%$ & $60\%$ & $65\%$ \\
\midrule

RS
& 6 & 14 & 187
& 15 & 155 & N/R \\

KCDS
& \textbf{5} & 7 & 18
& \textbf{7} & 15 & 20 \\

LCUS
& \textbf{5} & 7 & N/R
& \textbf{7} & 13 & N/R \\

$C_0$
& \textbf{5} & 7 & 11
& \textbf{7} & 9 & 14 \\

$C_{0.1}$
& \textbf{5} & 7 & 10
& \textbf{7} & 9 & 17 \\

$C_{0.5}$
& \textbf{5} & \textbf{6} & \textbf{9}
& \textbf{7} & \textbf{8} & \textbf{10} \\

\bottomrule
\end{tabular*}
\end{minipage}
\end{table}
\subsection{Heterogeneous Multi-Robot Online Adaptation}
\begin{figure}[htbp]
    \centering
    \includegraphics[width=0.7\columnwidth]
    {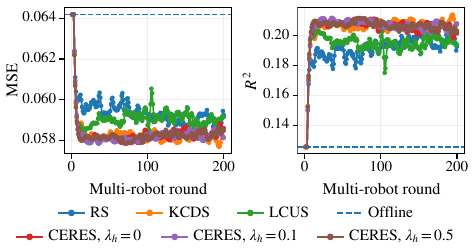}
    \caption{Prediction accuracy evaluated on held-out set during multi-robot adaptation: MSE (left) and $R^2$ (right).}
    \label{fig:performance_vs_queries}
\end{figure}

Table~\ref{tab:multi_robot_fixed_budget} reports prediction accuracy at selected batch rounds, and Fig.~\ref{fig:performance_vs_queries} shows its evolution over $200$ rounds. The dashed lines indicate the offline predictor.

The largest improvements occur early in adaptation. After $10$ rounds, CERES with $\lambda_h=0.5$ achieves an MSE of $0.0582$ and $R^2=0.2075$, compared with $(0.0601,0.1812)$ for RS, $(0.0588,0.1990)$ for KCDS, and $(0.0588,0.1995)$ for LCUS.
Relative to the offline predictor, this corresponds to an MSE reduction exceeding $9\%$ and a relative increase in $R^2$ of approximately $65.3\%$. The $\lambda_h=0$ and $0.1$ configurations also improve rapidly, reaching MSE values of $0.0585$ and $0.0583$.

CERES maintains prediction accuracy near the levels attained during these early rounds. The $\lambda_h=0.1$ reaches $(0.0579,0.2108)$ at round $20$, while $\lambda_h=0.5$ attains $(0.0579,0.2105)$ at round $100$. At every round in Table~\ref{tab:multi_robot_fixed_budget}, one CERES configuration achieves the lowest MSE and the highest $R^2$. The changing order of the CERES configurations shows that the benefit of uncertainty regularization depends on the stage of adaptation. Positive regularization provides additional gains over excitation-only sampling at
several reported rounds while preserving its strong initial prediction improvement.
\begin{table}[htbp]
\centering
\caption{Prediction performance during heterogeneous
multi-robot adaptation.
$C_{\lambda}$ denotes CERES with $\lambda_h=\lambda$.}
\label{tab:multi_robot_fixed_budget}

\scriptsize
\setlength{\tabcolsep}{0.5pt}
\renewcommand{\arraystretch}{0.94}

\begin{minipage}[t]{0.495\columnwidth}
\vspace{0pt}
\centering
\begin{tabular*}{\linewidth}{
@{\extracolsep{\fill}}c l c c c@{}
}
\toprule
Round & Method & MSE & $R^2$ & Cov. \\
\midrule
0 & Offline & 0.0642 & 0.1255 & -- \\
\midrule

\multirow{6}{*}{5}
& RS      & 0.0611 & 0.1675 & 0.7925 \\
& KCDS    & 0.0610 & 0.1684 & 0.7925 \\
& LCUS    & 0.0608 & 0.1715 & 0.7925 \\
& $C_0$   & 0.0609 & 0.1700 & 0.7950 \\
& $C_{0.1}$ & 0.0610 & 0.1695 & 0.7950 \\
& $C_{0.5}$ & \textbf{0.0607} & \textbf{0.1725} & 0.7950 \\
\addlinespace[2pt]

\multirow{6}{*}{10}
& RS      & 0.0601 & 0.1812 & 0.8000 \\
& KCDS    & 0.0588 & 0.1990 & 0.7950 \\
& LCUS    & 0.0588 & 0.1995 & 0.7950 \\
& $C_0$   & 0.0585 & 0.2032 & 0.7875 \\
& $C_{0.1}$ & 0.0583 & 0.2053 & 0.7900 \\
& $C_{0.5}$ & \textbf{0.0582} & \textbf{0.2075} & 0.7900 \\
\addlinespace[2pt]

\multirow{6}{*}{20}
& RS      & 0.0597 & 0.1864 & 0.7925 \\
& KCDS    & 0.0580 & 0.2092 & 0.8025 \\
& LCUS    & 0.0586 & 0.2016 & 0.7950 \\
& $C_0$   & 0.0581 & 0.2082 & 0.7900 \\
& $C_{0.1}$ & \textbf{0.0579} & \textbf{0.2108} & 0.8000 \\
& $C_{0.5}$ & 0.0582 & 0.2066 & 0.7975 \\

\bottomrule
\end{tabular*}
\end{minipage}%
\hfill%
\begin{minipage}[t]{0.495\columnwidth}
\vspace{0pt}
\centering
\begin{tabular*}{\linewidth}{
@{\extracolsep{\fill}}c l c c c@{}
}
\toprule
Round & Method & MSE & $R^2$ & Cov. \\
\midrule
0 & Offline & 0.0642 & 0.1255 & -- \\
\midrule

\multirow{6}{*}{60}
& RS      & 0.0595 & 0.1895 & 0.8025 \\
& KCDS    & 0.0584 & 0.2045 & 0.8000 \\
& LCUS    & 0.0589 & 0.1976 & 0.8025 \\
& $C_0$   & 0.0581 & 0.2085 & 0.7900 \\
& $C_{0.1}$ & \textbf{0.0579} & \textbf{0.2105} & 0.7950 \\
& $C_{0.5}$ & 0.0582 & 0.2073 & 0.8000 \\
\addlinespace[2pt]


\multirow{6}{*}{100}
& RS      & 0.0591 & 0.1945 & 0.8025 \\
& KCDS    & 0.0586 & 0.2017 & 0.8050 \\
& LCUS    & 0.0589 & 0.1974 & 0.8050 \\
& $C_0$   & 0.0581 & 0.2090 & 0.8025 \\
& $C_{0.1}$ & 0.0585 & 0.2031 & 0.7950 \\
& $C_{0.5}$ & \textbf{0.0579} & \textbf{0.2105} & 0.8075 \\
\addlinespace[2pt]

\multirow{6}{*}{200}
& RS      & 0.0592 & 0.1928 & 0.8075 \\
& KCDS    & 0.0587 & 0.2006 & 0.8100 \\
& LCUS    & 0.0592 & 0.1940 & 0.8100 \\
& $C_0$   & 0.0586 & 0.2015 & 0.8050 \\
& $C_{0.1}$ & \textbf{0.0581} & \textbf{0.2080} & 0.8050 \\
& $C_{0.5}$ & 0.0585 & 0.2032 & 0.8050 \\
\bottomrule
\end{tabular*}
\end{minipage}
\end{table}

\begin{figure}[htbp]
    \centering
    \includegraphics[width=0.7\columnwidth]
    {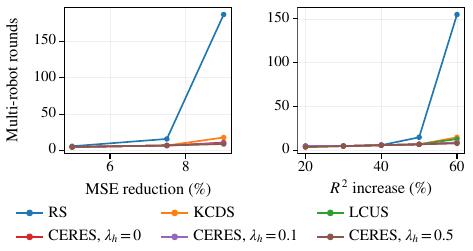}
    \caption{Multi-robot rounds required to attain relative MSE reductions (left) and $R^2$ increases (right).}
    \label{fig:sample_efficiency}
\end{figure}

The threshold comparisons in Table~\ref{tab:sample_efficiency}(b) and Fig.~\ref{fig:sample_efficiency} quantify the advantage during early adaptation. Each figure marker identifies the first round attaining the corresponding improvement target; targets not reached within the plotted horizon have no marker. CERES reaches a $9\%$ MSE reduction in $11$, $10$, and $9$ rounds for $\lambda_h=0$, $0.1$, and $0.5$, respectively. KCDS requires $18$ rounds and RS requires $187$, while LCUS does not attain this threshold within the plotted horizon. CERES with $\lambda_h=0.5$ therefore reduces the required rounds by $50\%$ relative to KCDS and $95.2\%$ relative to RS. The reduction from $11$ rounds for excitation only acquisition to $9$ rounds with $\lambda_h=0.5$ also shows the contribution of
uncertainty regularization to early prediction improvement.

A $60\%$ relative increase in $R^2$ is attained in $8$ rounds by CERES with $\lambda_h=0.5$, and in $9$ rounds by the other two CERES configurations. LCUS, KCDS, and RS require $13$, $15$, and $155$ rounds, respectively, as shown in Fig.~\ref{fig:sample_efficiency}. CERES with $\lambda_h=0.5$ requires the fewest evaluation rounds among the compared strategies for both prediction improvement targets. Together with the single robot results,
these findings demonstrate the evaluation efficiency of CERES under both sequential and heterogeneous batch evaluation.

\subsection{Empirical Coverage on Held-Out Evaluation Set}

Complementing the temporal calibration analysis in Lemma~\ref{prop:dtaci-local-calibration}, we assess coverage using the sequential held-out evaluation procedure in Section~\ref{sec:online_evaluation_metric}.
The coverage values in Tables~\ref{tab:single_robot_fixed_budget}
and~\ref{tab:multi_robot_fixed_budget} remain close to the target $1-\alpha=0.8$. They range from $0.7875$ to $0.8075$ in the single robot experiment and from $0.7875$ to $0.8100$ in the multi-robot experiment, giving a maximum deviation of $1.25$\% points from the target. Coverage is comparable across the sampling strategies despite their differences in
prediction accuracy and threshold achievement. CERES consequently combines the prediction improvements reported above with empirical coverage close to the prescribed rate in both settings.
\subsection{Excitation Uncertainty Tradeoff in Prompt Sampling}
\begin{figure}[htbp]
    \centering
    \includegraphics[width=0.7\columnwidth]
    {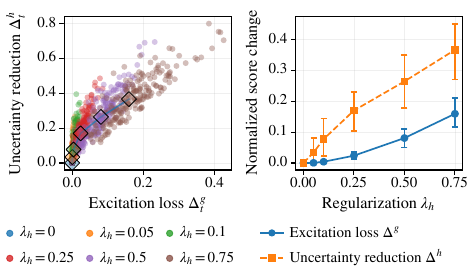}
    \caption{Excitation and uncertainty tradeoff: individual rounds (left) and median trends (right).}
    \label{fig:ceres_tradeoff}
\end{figure}

For objective~\ref{obj_3}, we examine the excitation and uncertainty tradeoff established in Theorem~\ref{thm:ceres_sampling_efficiency}. Figure~\ref{fig:ceres_tradeoff} evaluates the excitation uncertainty tradeoff for $\lambda_h$ between $0$ and $0.75$. Excitation loss and uncertainty reduction are measured relative to maximum excitation selection using the
normalized sampling scores. In the left panel, each point represents one round, and diamonds mark the medians of the two coordinates. The right panel shows the median trends, with error bars spanning the 25th--75th percentiles across rounds. Increasing $\lambda_h$ produces a greater median reduction in estimated local uncertainty and a greater median excitation loss. The variation across rounds reflects the changing prompt set and online estimates used for sampling. Thus, the penalty provides a systematic means of adjusting which informative prompts are prioritized for evaluation.

\begin{figure}[htbp]
    \centering
    \includegraphics[width=0.4\textwidth]
    {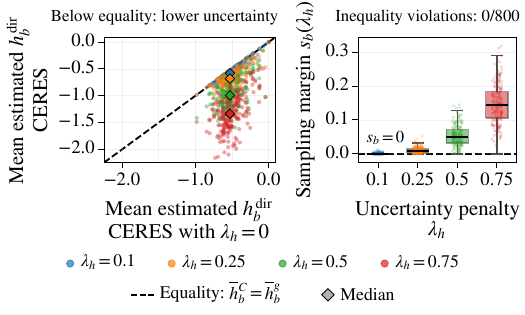}
    \caption{Paired batch comparisons of estimated uncertainty (left) and sampling margins $s_b(\lambda_h)$ (right).}
    \label{fig:theorem2_verification}
\end{figure}

Figure~\ref{fig:theorem2_verification} examines the corresponding tradeoff under batch selection by comparing CERES and excitation only batches formed using the same predictor, Gram matrix, candidate
prompt set, and local uncertainty estimates. The left panel compares batch means of the unnormalized scores $h_b^{\mathrm{dir}}$. The diagonal indicates equal means, and diamonds mark the medians of the two coordinates across rounds. Both batch sampling methods satisfy identical robot capacity constraints and have equal cardinality. The CERES batches have mean estimated uncertainty at or below that of the excitation-only batches. Larger values of $\lambda_h$ produce larger reductions, implying the penalty changes the sampled prompts even when all quantities available for ranking are held fixed.

For each paired comparison, $\Delta g_b$ and $\Delta h_b$
are the differences in mean normalized excitation and
uncertainty, respectively, obtained by subtracting the
CERES batch mean from the excitation-only batch mean.
Both batches use the same online state and normalization. The right panel shows the sampling margin $s_b(\lambda_h)\coloneqq\lambda_h\Delta h_b-\Delta g_b$,
with the dashed line marking zero. No violations of the nonnegative margin condition are reported among the $800$ comparisons. The observed excitation loss is therefore accompanied by the corresponding reduction in estimated uncertainty prescribed by the sampling criterion.

The prediction and sampling results together explain the role of uncertainty regularization within CERES. Feature excitation promotes evaluation along poorly represented feature directions, while the local statistic accounts for residual variation after bias compensation. Their combination supports earlier prediction improvement and changes prompt sampling in the intended direction as the uncertainty weight increases.

\subsection{Local Uncertainty and Observed Prediction Error}
\begin{figure}[htbp]
    \centering
    \includegraphics[width=0.6\columnwidth]
    {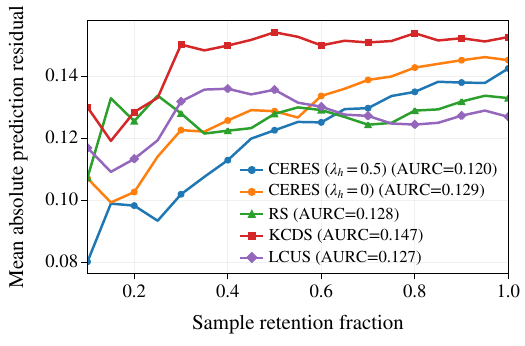}
    \caption{Selective prediction risk on each method's sampled prompts.}
    \label{fig:selective_risk}
\end{figure}

For objective~\ref{obj_2}, we examine whether the local uncertainty statistic supported by Lemma~\ref{prop:local_knn_coverage} and
Proposition~\ref{prop:local_conformal_entropy_bound} identifies queried prompts with smaller prediction errors. Figure~\ref{fig:selective_risk} orders each method's evaluated prompts by increasing uncertainty recorded at sampling time. At each sample fraction, the curve reports the mean absolute pre-update prediction residual among the prompts with the lowest uncertainty scores. The curves therefore evaluate the relationship between uncertainty and observed error along the prompt sequences acquired by the respective methods. The legend reports the area under each curve over sample fractions $[0.1,1]$, divided by the interval length $0.9$.

CERES with $\lambda_h=0.5$ achieves the lowest area under the curve, $0.120$, compared with $0.129$ for maximum excitation sampling, $0.128$ for RS, $0.147$ for KCDS, and $0.127$ for LCUS. This corresponds to a $6.98\%$ reduction relative to
maximum excitation sampling. Its advantage is most pronounced among the smallest sample fractions, where the prompts assigned the lowest uncertainty exhibit smaller observed prediction errors. As the fraction increases, the curves become closer and their ordering changes. The integrated comparison favors CERES with $\lambda_h=0.5$ over the plotted range of fractions.

This result complements the paired sampling comparison. Uncertainty regularization favors batches with lower estimated local uncertainty, and the resulting uncertainty rankings identify subsets with lower observed prediction error. Combined with the evaluation counts and held-out set accuracy measurements, the results
support the use of CERES to allocate costly controller evaluations while improving the assessment of unevaluated prompts.

\section{Conclusion}
\label{sec:conclusion}
This work presented CERES for improving the prediction of LLM-guided MPC performance under a limited closed-loop evaluation budget. CERES integrates online least squares adaptation and temporal calibration with prompt sampling that balances feature excitation and bias-compensated local uncertainty. Our analysis identifies conditions for a tighter expected static prediction regret bound and establishes local coverage, conditional differential entropy, and sequential feature coverage bounds. Experiments demonstrate earlier achievement of prescribed prediction accuracy targets than the compared baselines in both single-robot and heterogeneous multi-robot settings, with held-out coverage near the nominal level. 


\appendix
\subsubsection{\textbf{Performance Score Details}}
\label{app:score}

For task $I$, consider trajectories $X=(x_0,\ldots,x_{N_I})$ and $U=(u_0,\ldots,u_{N_I-1})$, with $N_I\geq1$. The four components of Eq.~\eqref{eq:prelim-performance-score} are objective completeness, arrival time, control effort, and control smoothness and each lies in $[0,1]$.

The minimum target violation along the trajectory is defined as $m_{s}(X;I)=\min_{0\leq k\leq N_I}\phi_{T,I}(x_k)$, and the objective completeness score is $S_{\mathrm{obj}}(X,U;I)=\tfrac{1}{1+m_+(X;I)/D_I}$, 
where $m_+(X;I)=\max\{m_{s}(X;I),0\}$ and $D_I>0$ has the same units as $\phi_{T,I}$. The score equals one when the target is reached and decreases with the smallest positive violation.

The first target entry index is $k_{\mathrm{hit}}=\min\{k:\phi_{T,I}(x_k)\leq0\}$ if the target is reached, and $k_{\mathrm{hit}}=N_I$ otherwise. 
The arrival time score is $S_{\mathrm{time}}(X,U;I)=1-(k_{\mathrm{hit}}/N_I)^{2}$, assigning larger values to earlier arrival and zero to failures.

For input dimension $m$ and bound $\|u_k\|_\infty\leq u_{m}$.
The effort and smoothness scores are $e_u(U;I)=\tfrac{\sum_{k=0}^{N_I-1}\|u_k\|_2^2}{N_I m u_{m}^2}$ and $e_{\Delta u}(U;I)=\tfrac{\sum_{k=1}^{N_I-1}\|u_k-u_{k-1}\|_2^2}{N_I(\Delta u_{\mathrm{ref}})^2}$
with $\Delta u_{\mathrm{ref}}=0.25u_{m}$. Their scores are $S_{\mathrm{eff}}=1-\Pi_{[0,1]}(e_u)$ and $S_{\mathrm{smth}}=1-\Pi_{[0,1]}(e_{\Delta u})$. The effort denominator corresponds to maximum input magnitude in every dimension throughout the horizon. Eq.~\eqref{eq:prelim-performance-score} combines these components using the preference weights $\rho$.
\subsubsection{\textbf{Structured Prompt Design}}
\label{app:prompt-example}

The task descriptors in Section~\ref{subsubsec:prompt-construction} are defined below. The output specification fixes the parsing interface used throughout controller evaluation.

\noindent\emph{\underline{Task Characteristic Descriptors}}.
\label{app:descriptor-definitions}
The normalized initial distance is $d_0^{\mathrm{norm}}=\|x_0-x_{\mathrm{tar}}\|_2/r$. With processed linear drift matrix $A$ and constant input map $G_0$, the linear drift descriptor is $\varrho_{\mathrm{lin}}=\max_i|\lambda_i(I+\Delta t A)|$, and the nominal input gain is $g_{\mathrm{ctrl}}=\|G_0\|_2$. The nonlinear dynamics descriptor is $\eta_{\mathrm{nl}}= \sum_{q=2}^{d_f}\|C^{(q)}\|_{\mathrm F}r^{q-1}+u_{\max}\sum_{q=1}^{d_G}\|D^{(q)}\|_{\mathrm F}r^{q-1}$, 
where $\|\cdot\|_{\mathrm F}$ is the square root of the sum of squared array entries, including all three indices of $D^{(q)}$. This quantity describes nonlinear magnitude and is not a Lipschitz or Jacobian bound. Finally, $r_{\mathrm{reach}}=\|x_0-x_{\mathrm{tar}}\|_2/(N_I\Delta t\,u_{\max}g_{\mathrm{ctrl}})$ compares target displacement with the evaluation time, admissible input magnitude, and nominal input gain.

\noindent\emph{\underline{Output Specification}}.
\label{app:representative-prompt}
The prompt combines the system instruction, task descriptor, and preference profile. The response must be a single JSON object containing only the key \texttt{w}, with seven finite nonnegative weights in the order specified in Section~\ref{subsubsec:llm-weight-generation}. Additional keys or text are invalid.
\subsubsection{\textbf{Proof of Lemma~\ref{prop:dtaci-local-calibration}}}
\label{app:proof_dtaci_local_calibration}
For the projected experts in Section~\ref{sec:adaptive_conformal_preliminaries}, nonexpansiveness of projection onto $[0,1]$ and the unit bound on the pinball-loss subgradients give
\begin{equation}
    \sum_{t\in\mathcal T}
    \mathbb E\!\left[
        \ell_\alpha(\alpha_t^{(j)};\beta_t)
        -
        \ell_\alpha(\alpha_t^\star;\beta_t)
    \right]
    \leq
    \tfrac{1+2V_{\mathcal T}^{\alpha}}{2\gamma_j}
    +\tfrac{\gamma_j|\mathcal T|}{2}.
\end{equation}
Combining this projected expert bound with the DtACI aggregation argument and the density based loss comparison in~\cite[Sec.~3.1 and Prop.~5]{JMLR:v25:22-1218} yields Eq.~\eqref{eq:dtaci-local-alpha-bound} under the stated gain grid and parameter conditions.

By conditional independence of the DtACI expert sampling, the oracle condition, and the Lipschitz coverage map, we obtain, for each $t\in\mathcal T$,
\begin{equation}
\begin{aligned}
    &|
        \mathbb P (
            Y_t^{\mathrm{LLM}}\in\widehat C_t(\alpha_t)
        )-(1-\alpha)
    |\\
    &=
    |
        \mathbb E [
            c_t^{\mathrm{cov}}(\alpha_t;\mathcal H_{t-1})
            -
            c_t^{\mathrm{cov}}(\alpha_t^\star;\mathcal H_{t-1})
        ]
    |\\
    &\leq
    L_{\mathrm{cov}}\,
    \mathbb E [
        |\alpha_t-\alpha_t^\star|
    ].
\end{aligned}
\label{eq:conditional-coverage-step}
\end{equation}

By Cauchy-Schwarz inequalities and Eq.~\eqref{eq:dtaci-local-alpha-bound}, summing Eq.~\eqref{eq:conditional-coverage-step} gives $\sum_{t\in\mathcal T}
    |
        \mathbb P( Y_t^{\mathrm{LLM}}\in\widehat C_t(\alpha_t) )-(1-\alpha)
    \leq
    L_{\mathrm{cov}}
    \sqrt{
        |\mathcal T|
        \sum_{t\in\mathcal T}
        \mathbb E[
            (\alpha_t-\alpha_t^\star)^2
        ]
    }\leq
    |\mathcal T|L_{\mathrm{cov}}
    \sqrt{\tfrac{2B_{\mathcal T}}{p_\beta}}
    =
    |\mathcal T|\varepsilon_{\mathcal T}$.
This proves Eq.~\eqref{eq:dtaci-summed-miscoverage} by taking complements.
\subsubsection{\textbf{Proof of Theorem~\ref{thm:online-ls-regret}}}
\label{app:proof-online-ls-regret}
The standard projected online gradient inequality~\cite[Sec.~3.1]{hazan2016introduction}, applied with constant learning rate $\eta$ and comparator $\omega^\star\in\Omega$, gives
\begin{equation}
 \begin{aligned}
 \operatorname{Regret}_T
 &\leq\tfrac{R_\Omega^2}{2\eta}
 +\tfrac{\eta}{2}\sum_{t=1}^T
 \|\nabla_\omega\ell_t(\widehat\omega_t)\|_2^2\\
 &\leq\tfrac{R_\Omega^2}{2\eta}+2\eta L_\Phi^2\sum_{t=1}^T\xi_t^2.
 \end{aligned}
 \label{eq:regret-gradient-energy-proof}
\end{equation}
where the second inequality uses $\nabla_\omega\ell_t(\widehat\omega_t)=-2\Phi_t\xi_t$ and $\|\Phi_t\|_2\leq L_\Phi$. 
The uniform bound in Eq.~\eqref{eq:uniform_prediction_residual_bound} and the calibrated bound in Eq.~\eqref{eq:calibrated_residual_energy_bound} together give $\sum_{t=1}^T\mathbb E[\xi_t^2]\leq TM_\xi^2\vartheta_T$ by Eq.~\eqref{eq:calibrated_residual_factor}. Taking expectations in Eq.~\eqref{eq:regret-gradient-energy-proof} yields Eq.~\eqref{eq:expected-regret-final-theorem}. For $R_\Omega>0$, setting $\eta=R_\Omega/(2L_\Phi M_\xi\sqrt T)$ gives Eq.~\eqref{eq:expected-regret-refined-rate-theorem}.

\subsubsection{\textbf{Proof of Lemma~\ref{lem:directional_residual_separation}}}
\label{app:proof_directional_residual_separation}
For $i\in\mathcal N_t(p)$, the decomposition is
$\zeta_{i,t}=b_t(p_i)+\varepsilon_i$, where
$\varepsilon_i=Y_i^{\mathrm{LLM}}-m_\Phi(p_i)$.
The Lipschitz gradient assumption gives
$b_t(p_i)=b_t(p)+\nabla_{\bar\Phi}b_t(p)^\top d_i(p)+r_{i,t}(p)$,
with $|r_{i,t}(p)|\leq L_b\|d_i(p)\|_2^2/2$.
By the reverse triangle inequality and $\|d_i(p)\|_2\leq\rho_t(p)$, we have
$\left|u_{i,t}^{\mathrm{dir}}(p)-|\varepsilon_i|\right| \leq |b_t(p)-\widehat b_t(p)|
+\rho_t(p)\|\nabla_{\bar\Phi}b_t(p) -\widehat{\boldsymbol\beta}_t(p)\|_2 +\tfrac{L_b}{2}\rho_t(p)^2 \leq\varepsilon_{\mathrm{reg},t}(p) +\tfrac{L_b}{2}\rho_t(p)^2=\delta_t^{\mathrm{dir}}(p)$, 
where the last inequality uses Eq.~\eqref{eq:local_affine_fit_error} and the claimed bound therefore holds for all $i\in\mathcal N_t(p)$.
\subsubsection{\textbf{Proof of Lemma~\ref{lem:local_quantile_error}}}
\label{app:proof_local_quantile_error}
Under the indexing in Assumption~\ref{ass:local_sampling_model}, $\mathcal R_t$ contains the first $N_0+t-1$ observations.
Settin $A=\log(2N_{\mathrm{pool}}^3/\delta_{\mathrm{conf}})$
and $\kappa_n=\sqrt{A/(2n)}$ gives
\begin{equation}
 \underline f\,\varepsilon_{\mathrm{stat}}(n)=\kappa_n+1/n.
 \label{eq:proof_quantile_error_scale}
\end{equation}
The noise stationarity makes $q_p=q_{\varepsilon,t}^{\star}(p;\tau)$ independent of $t$.

\noindent\emph{\underline{Distance prefixes and probability bounds}}.
For each $p\in\mathcal P$, order the pool by distance from $\bar\Phi(p)$, resolving ties by the fixed prompt ordering. Let $B$ be a nonempty prefix and $\rho_B(p)=\max_{p'\in B}\|\bar\Phi(p')-\bar\Phi(p)\|_2$. Each realized neighborhood consists of the evaluated members of one such prefix. Fix $n\in\{1,\ldots,|B|\}$ satisfying $\varepsilon_{\mathrm{stat}}(n)\leq r_q$, and set $z^\pm=q_p\pm[L_q\rho_B(p)+\varepsilon_{\mathrm{stat}}(n)]$. For every $p'\in B$, quantile smoothness gives $|q_{p'}-q_p|\leq L_q\rho_B(p)$. Integrating the density lower bound within $\varepsilon_{\mathrm{stat}}(n)$ of $q_{p'}$ and using CDF monotonicity yields
\begin{equation}
\begin{aligned}
 F_{p'}^\varepsilon(z^-)&\leq1-\tau-\underline f\,\varepsilon_{\mathrm{stat}}(n),\\
 F_{p'}^\varepsilon(z^+)&\geq1-\tau+\underline f\,\varepsilon_{\mathrm{stat}}(n),
\end{aligned}
\label{eq:proof_noise_threshold_probabilities}
\end{equation}
where the density assumption also ensures $F_{p'}^\varepsilon(q_{p'})=1-\tau$.

\noindent\emph{\underline{Concentration under adaptive selection}}.
Define $\chi_i(B)=\mathbf1\{p_i\in B\}$, $N_j(B)=\sum_{i=1}^j\chi_i(B)$, and
\begin{equation}
 S_j(z)=\sum_{i=1}^j\chi_i(B)
 \bigl[\mathbf1\{|\varepsilon_i|\leq z\}-F_{p_i}^\varepsilon(z)\bigr].
\end{equation}
Conditional on $\mathcal F_{i-1}$ and $p_i$, each increment has mean zero by Eq.~\eqref{eq:local_noise_stability} and range length $\chi_i(B)$. Conditional Hoeffding's lemma, followed by averaging over prompt selection, therefore makes
$M_j(z,\lambda)=\exp\{\lambda S_j(z)-\lambda^2N_j(B)/8\}$ a nonnegative $\mathcal F_j$-supermartingale with $M_0=1$.
When $N_j(B)=n$ and $S_j(z^-)\geq n\kappa_n$, the choice $\lambda=4\kappa_n$ gives $\log M_j(z^-,\lambda)\geq2n\kappa_n^2=A$. Ville's inequality and the analogous choice $\lambda=-4\kappa_n$ at $z^+$ imply
$\mathbb P\bigl(\exists j:N_j(B)=n,\ S_j(z^-)\geq n\kappa_n\bigr)\leq e^{-A}$ and $\mathbb P\bigl(\exists j:N_j(B)=n,\ S_j(z^+)\leq-n\kappa_n\bigr)\leq e^{-A}$.

\noindent\emph{\underline{Quantile inversion and uniform control}}.
At $N_j(B)=n$, let $\widehat F_{j,B}$ and $q_{j,B}^{\mathrm{emp}}$ be the empirical CDF and $(1-\tau)$-quantile of the selected noise magnitudes in $B$.
The rank $\lceil n(1-\tau)\rceil$ convention gives
$q_{j,B}^{\mathrm{emp}}<z^-\Rightarrow\widehat F_{j,B}(z^-)\geq1-\tau-1/n$
and
$q_{j,B}^{\mathrm{emp}}>z^+\Rightarrow\widehat F_{j,B}(z^+)\leq1-\tau+1/n$.
These implications also hold for linear interpolation at rank $1+(n-1)(1-\tau)$; the $1/n$ allowance covers adjacent order statistics. Since $S_j(z)/n=\widehat F_{j,B}(z)-n^{-1}\sum_{i=1}^j\chi_i(B)F_{p_i}^\varepsilon(z)$, Eqs.~\eqref{eq:proof_quantile_error_scale} and~\eqref{eq:proof_noise_threshold_probabilities} give $q_{j,B}^{\mathrm{emp}}<z^-\ \Longrightarrow\ S_j(z^-)\geq n\kappa_n$ and $q_{j,B}^{\mathrm{emp}}>z^+\ \Longrightarrow\ S_j(z^+)\leq-n\kappa_n$.
Thus, for each fixed $(p,B,n)$, the probability of a quantile outside $[z^-,z^+]$ at any such evaluation index is at most $2e^{-A}$. There are at most $N_{\mathrm{pool}}$ choices each of $p$, its prefix $B$, and the positive count $n$, because prompts are evaluated at most once. A union bound gives an event $\mathcal E_{\mathrm{conf}}$ of probability at least $1-2N_{\mathrm{pool}}^3e^{-A}=1-\delta_{\mathrm{conf}}$ on which all these bounds hold.

For any admissible round $t$ and $p\in\mathcal C_t$, choose $B$ ending at the $K_t$-th nearest prompt in $\mathcal R_t$. Its evaluated members are exactly $\mathcal N_t(p)$, so $N_{N_0+t-1}(B)=K_t$, $\rho_B(p)=\rho_t(p)$, and $q_{N_0+t-1,B}^{\mathrm{emp}}=q_{\varepsilon,t}^{\mathrm{emp}}(p;\tau)$. The bound for $(p,B,K_t)$ establishes Eq.~\eqref{eq:local_noise_quantile_concentration} simultaneously for all candidates and rounds with $K_t\geq1$ and $\varepsilon_{\mathrm{stat}}(K_t)\leq r_q$.
\subsubsection{\textbf{Proof of Lemma~\ref{prop:local_knn_coverage}}}
\label{app:proof_local_knn_coverage}
Given $p,t$ satisfying the lemma and consider $\mathcal E_{\mathrm{conf}}$. Lemma~\ref{lem:directional_residual_separation} bounds each score perturbation by $\delta_t^{\mathrm{dir}}(p)$. The same bound applies to corresponding order statistics and their empirical quantiles. Combining it with Lemma~\ref{lem:local_quantile_error} gives $|\widehat q_t^{\mathrm{dir}}(p;\tau)-q_{\varepsilon,t}^{\star}(p;\tau)|\leq\delta_t^{\mathrm{dir}}(p)+L_q\rho_t(p) +\varepsilon_{\mathrm{stat}}(K_t)$. 
Projection onto $[0,1]$ cannot increase the distance to $m_\Phi(p)\in[0,1]$. Thus Eq.~\eqref{eq:local_affine_fit_error} and
Eq.~\eqref{eq:directional_local_center} imply
$|\widehat m_t^{\mathrm{dir}}(p)-m_\Phi(p)|
\leq|\widehat b_t(p)-b_t(p)|\leq\varepsilon_{\mathrm{reg},t}(p)$.
Using Eq.~\eqref{eq:directional_coverage_perturbation}, whenever $|\varepsilon_t(p)|\leq q_{\varepsilon,t}^{\star}(p;\tau) -\Delta_t^{\mathrm{dir}}(p)$, the triangle inequality gives $|Y_t^{\mathrm{LLM}}(p)-\widehat m_t^{\mathrm{dir}}(p)| \leq |\varepsilon_t(p)|+\varepsilon_{\mathrm{reg},t}(p)
 \leq\widehat q_t^{\mathrm{dir}}(p;\tau)$.
Since $Y_t^{\mathrm{LLM}}(p)\in[0,1]$, intersection with $[0,1]$
preserves this inclusion in $\widehat C_t^{\mathrm{dir}}(p;\tau)$.
Consequently, Assumption~\ref{ass:local_cdf_regularity} yields $\mathbb P (Y_t^{\mathrm{LLM}}(p)\in \widehat C_t^{\mathrm{dir}}(p;\tau)\mid\mathcal H_{t-1})\geq F_{t,p}^{\varepsilon} (q_{\varepsilon,t}^{\star}(p;\tau)-\Delta_t^{\mathrm{dir}}(p)) \geq 1-\tau-c\Delta_t^{\mathrm{dir}}(p) =1-\tau-\varepsilon_t^{\mathrm{dir}}(p)$,
where the last step uses the oracle quantile and
$\varepsilon_t^{\mathrm{dir}}(p)=c\Delta_t^{\mathrm{dir}}(p)$.
\subsubsection{\textbf{Proof of Proposition~\ref{prop:local_conformal_entropy_bound}}}
\label{app:proof_local_entropy_bound}
For fixed $p,t$ satisfying the proposition, denote
$Y=Y_t^{\mathrm{LLM}}(p)$, $C=\widehat C_t^{\mathrm{dir}}(p;\tau)$,
and $V=V_t^{\mathrm{dir}}(p)$. Let $E=\mathbf1\{Y\in C\}$ and
$\pi=\mathbb P(E=1\mid\mathcal H_{t-1})$.
On $\mathcal E_{\mathrm{conf}}$, Lemma~\ref{prop:local_knn_coverage}
and Eq.~\eqref{eq:uniform_local_eta_definition} give
$1-\pi\leq\bar\eta_{\mathrm{dir}}$.
The bound is immediate for $H(Y\mid\mathcal H_{t-1})=-\infty$.
For $V=1$, it follows from $H(Y\mid\mathcal H_{t-1})\leq0$
and $h_b(\bar\eta_{\mathrm{dir}})\geq0$.
Consider finite conditional entropy and $0<V<1$.

Given $\mathcal H_{t-1}$, the interval is fixed and $E$ is determined by $Y$. Comparison with the uniform density on $C$ gives
$H(Y\mid\mathcal H_{t-1},E=1)\leq\log V$ by nonnegativity of
Kullback-Leibler divergence. The same argument on the complement gives $H(Y\mid\mathcal H_{t-1},E=0)\leq\log(1-V)$. The entropy chain rule therefore yields $H(Y\mid\mathcal H_{t-1}) =h_b(\pi)+\pi H(Y\mid\mathcal H_{t-1},E=1)
 +(1-\pi)H(Y\mid\mathcal H_{t-1},E=0)
 \leq h_b(\pi)+\pi\log V+(1-\pi)\log(1-V)
 \leq h_b(1-\pi)+\pi\log V$.
Zero probability events contribute zero, with $0\log0=0$. The last inequality uses $\log(1-V)\leq0$ and symmetry of $h_b$.
For $\psi(u)=h_b(u)+(1-u)\log V$,
$\psi'(u)=\log((1-u)/u)-\log V>0$ on $(0,1/2)$;
continuity at zero extends monotonicity to $[0,1/2)$.
Applying $1-\pi\leq\bar\eta_{\mathrm{dir}}<1/2$ and
$h_t^{\mathrm{dir}}(p)=\log V$ proves
Eq.~\eqref{eq:local_conformal_expected_volume_bound}.
\subsubsection{\textbf{Proof of Theorem~\ref{thm:ceres_sampling_efficiency}}}
\label{app:proof_ceres_sampling_efficiency}
Optimality in Eq.~\eqref{eq:state_dependent_sampler} gives $\Delta g_t\leq\lambda_h\Delta h_t$. Since $p_t^g$ maximizes excitation and the normalized uncertainty scores lie in $[0,1]$, we obtain $0\leq\Delta g_t\leq\lambda_h\Delta h_t\leq\lambda_h$. Hence $\Delta h_t\geq0$, which implies $h_t^{\mathrm{dir}}(p_t^{\mathrm C})\leq h_t^{\mathrm{dir}}(p_t^g)$, including constant uncertainty scores.


If the excitation scores are not identical across $\mathcal C_t$, then $\widetilde g_t(p_t^g)=1$. The normalization and $\Delta g_t\leq\lambda_h$ give $g_t(p_t^{\mathrm C}) \geq (1-\lambda_h)m_t+\lambda_h g_t^{m} \geq (1-\lambda_h)m_t$. If all prompts in $\mathcal C_t$ have the same excitation
score, then $g_t(p_t^{\mathrm C})=m_t$, and the bound also holds.

Since $\Sigma_{t+1}^{-1}\preceq\Sigma_t^{-1}$ and $\mathcal C_{t+1}\subseteq\mathcal C_t$, we have $m_{t+1}\leq m_t$. Consequently, $(1-\lambda_h)T m_{T+1} \leq(1-\lambda_h)\sum_{t=1}^T m_t \leq\sum_{t=1}^T g_t(p_t^{\mathrm C}).$
The assumptions give $0\leq g_t(p_t^{\mathrm C})\leq G$. Using $x\leq(1+G)\log(1+x)$ for $x\in[0,G]$ and the matrix determinant lemma yields $\sum_{t=1}^T g_t(p_t^{\mathrm C}) \leq(1+G)\log\tfrac{\det\Sigma_{T+1}}{\det\Sigma_1} \leq(1+G)(d_\phi+1)\log(1+TG/(d_\phi+1))$. The final inequality follows from the arithmetic geometric mean inequality applied to the eigenvalues of $\Sigma_1^{-1/2}\Sigma_{T+1}\Sigma_1^{-1/2}$.
Their sum is at most $d_\phi+1+TG$, since each evaluation
contributes $\Phi(p_t^{\mathrm C})^\top\Sigma_1^{-1}\Phi(p_t^{\mathrm C})
\leq G$ to the trace. Combining the last two inequalities proves Eq.~\eqref{eq:ceres_maximum_excitation_rate}.
\subsubsection{\textbf{Proof of Corollary~\ref{cor:ceres_oracle_uncertainty}}}
\label{app:proof_ceres_oracle_uncertainty}
For $p,t$ satisfying the corollary, Appendix~\ref{app:proof_local_knn_coverage}
establishes, on $\mathcal E_{\mathrm{conf}}$,
$|\widehat m_t^{\mathrm{dir}}(p)-m_\Phi(p)|\leq\varepsilon_{\mathrm{reg},t}(p)$
and the quantile error bound
$|\widehat q_t^{\mathrm{dir}}(p;\tau)-q_{\varepsilon,t}^{\star}(p;\tau)|
 \leq\delta_t^{\mathrm{dir}}(p)+L_q\rho_t(p)
       +\varepsilon_{\mathrm{stat}}(K_t)$.
The width of $[m-q,m+q]\cap[0,1]$ is
$\min\{1,m+q\}-\max\{0,m-q\}$ for $m\in[0,1]$ and $q\geq0$.
Since both endpoint operations are $1$-Lipschitz, these bounds and
Eq.~\eqref{eq:directional_coverage_perturbation} imply
$|V_t^{\mathrm{dir}}(p)-V_{\varepsilon,t}^{\star}(p)|
\leq2\Delta_t^{\mathrm{dir}}(p)$.
For positive widths $x,y$, the mean value theorem gives $|\log x-\log y|\leq |x-y|/\min\{x,y\}$. Applying this inequality to the two widths proves Eq.~\eqref{eq:directional_uncertainty_approximation}. Applying that estimate at $p_t^{\mathrm C}$ and $p_t^g$ to the ordering in Eq.~\eqref{eq:ceres_excitation_uncertainty_consequences} gives Eq.~\eqref{eq:ceres_oracle_uncertainty_comparison}.

\bibliographystyle{IEEEtran}
\bibliography{sihangbib_shortened}

\end{document}